\documentclass[journal]{IEEEtran}
\usepackage{graphicx}
\usepackage{amsfonts}
\usepackage{amsmath}
\usepackage{amssymb}
\usepackage{amsfonts}
\usepackage{multirow}
\usepackage{makecell}
\usepackage{booktabs}
\usepackage[caption=false,font=tiny]{subfig}
\usepackage{placeins}

\ifCLASSINFOpdf
\else
\fi
\begin{document}

\title{Driver Gaze Target Identification via Gaze Estimation and Structured Driving Scene Understanding}
%
%
%

\author{Xinze~Cao,~\IEEEmembership{}
        Ke~Wang,~\IEEEmembership{Member,~IEEE,}
        Deyong~Guan,~\IEEEmembership{Member,~IEEE,}
        Yuanquan~Li,~\IEEEmembership{}
        and~Ke~Zhang~\IEEEmembership{}
\thanks{Corresponding author: Ke Wang.}
\thanks{Xinze Cao, Ke Wang, Deyong Guan, Yuanquan Li and Ke Zhang are with the College of Transportation, Shandong University of Science and Technology,
Qingdao 266590, China (e-mail: xinzecao@sdust.edu.cn; wangke@sdust.edu.cn)}
\thanks{}}

%
%

\markboth{Driver Gaze Target Identification via Gaze Estimation and Structured Driving Scene Understanding}%
{Cao \MakeLowercase{\textit{et al.}}: Driver Gaze Target Identification}
%



\maketitle

\begin{abstract}
Identifying a driver's gaze target provides object-level information for visual attention monitoring in intelligent vehicles. Gaze-zone classification and continuous attention localization provide limited information about the identity of the attended scene target. This paper proposes a modular framework that identifies object- or region-level gaze targets from synchronized in-cabin driver images and forward-view road images. The framework comprises four components: MFGaze, which combines a Vision Transformer pretrained with a masked autoencoder (MAE) and head-pose conditioning through feature-wise linear modulation (FiLM); panoptic scene understanding, which constructs a structured candidate set; polynomial ridge regression, which maps gaze features to 2D scene-image coordinates through individualized calibration; and entropy-driven target inference, which adaptively fuses boundary-distance geometry with class-level semantic priors. Experiments evaluate the framework using three public gaze datasets, a laboratory test using Cityscapes road scenes and self-collected gaze data, and the Look Both Ways (LBW) dataset. The complete framework achieves an overall Top-1 gaze target identification accuracy of 96.27\% in the laboratory experiment. After sequence-specific calibration, it achieves Top-1 accuracies of 79.92\% for task-relevant objects and 85.58\% for the road class on LBW. MFGaze achieves the lowest mean angular errors among the compared methods, with 4.06$^\circ$, 4.23$^\circ$, and 10.52$^\circ$ on MPIIFaceGaze, ETH-XGaze, and Gaze360, respectively, and demonstrates strong cross-dataset generalization. Sensitivity analyses show stable performance under hyperparameter perturbations. These results support the framework as a source of object-level gaze information for driver monitoring.
\end{abstract}


%
\IEEEpeerreviewmaketitle

\newpage

\section{Introduction}
\IEEEPARstart{D}{river} visual attention is a key indicator of driver state and plays an important role in advanced driver-assistance systems (ADAS) and intelligent vehicles~\cite{1}. Large-scale naturalistic driving studies indicate that approximately 80\% of crashes and 65\% of near-crashes involve driver inattention within 3 seconds prior to the event~\cite{2}. Understanding where and what the driver is attending to is not only a perception problem but also a safety-critical issue in driving systems~\cite{3}. Although autonomous driving technologies have advanced rapidly, human drivers remain essential in lower levels of driving automation and human–machine cooperation~\cite{4}. In this context, gaze-related driver attention analysis has broad safety-critical application potential, including distracted-driving risk assessment~\cite{5}, driver attention monitoring and takeover-related assessment~\cite{6}, and driving-intention prediction~\cite{7}. Object-level gaze-target information can therefore support driver state monitoring (DSM) by indicating which traffic objects or scene regions receive the driver's visual attention.

Identifying the object of driver attention in complex traffic environments is a fundamental yet challenging problem. Driver gaze and the surrounding traffic scene are observed from two different viewpoints: an in-cabin driver-facing camera and an out-of-cabin forward-facing camera. Associating gaze with scene targets therefore requires mapping gaze features into the coordinate space of the forward-view image. Gaze estimation and calibration errors may displace the mapped point of gaze from the attended target~\cite{8,9}. Meanwhile, densely distributed traffic participants and large differences in target size introduce ambiguity among neighboring candidates. These challenges motivate a calibrated gaze-to-scene mapping and an association mechanism that combines spatial proximity with scene semantics.

Existing studies have progressively improved the granularity of driver attention modeling. One line of research formulates driver attention estimation as gaze-zone classification, in which the cockpit or the driver’s field of view is divided into a predefined set of regions. Representative studies include feature-based six-region classification~\cite{10}, multi-view head-pose-based eight-zone inference~\cite{11}, and learning-based multi-channel networks~\cite{12}. A second line of research focuses on continuous attention localization, where driver attention is represented as 2D point-of-gaze (PoG) coordinates or attention heatmaps on the forward-view scene image. Recent methods further enhance this formulation by incorporating scene saliency~\cite{13}, driving-task priors~\cite{14}, or end-to-end cross-view fusion of in-cabin gaze and out-of-cabin scene features~\cite{15}. Despite recent progress, there remains a lack of low-cost dual-camera solutions that can directly identify object-level gaze targets in complex driving scenes.

This paper formulates driver gaze target identification as an object-level recognition problem and addresses the above challenges through a modular framework that decouples gaze estimation, scene understanding, gaze-to-scene calibration and mapping, and gaze target inference. The framework adopts a modular design to support interpretable and flexible gaze target identification. Each subproblem is defined with a clear input–output interface and can be evaluated independently. Individual components can be replaced, compressed, or optimized separately. In addition, task-specific inductive priors, such as drivers’ stronger attention to safety-relevant traffic participants, can be incorporated at the inference stage where semantic information is most useful. Given a synchronized pair of driver-facing and forward-facing images, the framework outputs the driver’s current gaze target for each frame.

The framework comprises four cooperating modules. First, the MFGaze module estimates gaze direction using a self-supervised Vision Transformer backbone with explicit head-pose conditioning, improving robustness to illumination changes, partial occlusions, and large head rotations. Second, the scene understanding module transforms the panoptic segmentation of the forward-view image into a structured candidate object set, in which traffic participants and background regions are represented with semantic and geometric attributes. Third, the calibration module maps gaze features to 2D scene-image coordinates through polynomial ridge regression, compensating for individual posture differences and residual gaze estimation errors without external depth. Finally, the target inference module fuses boundary-distance geometry with class-level semantic priors and adaptively adjusts their contributions according to scene ambiguity through an entropy-driven mechanism.

The main contributions of this paper are summarized as follows.

1)  We formulate driver gaze target identification as an object-level recognition task and propose a modular framework that maps the driver's gaze from raw facial images to the attended target in the forward-view scene for driver visual attention monitoring, providing object-level outputs that previous gaze-zone classification and continuous spatial attention localization methods do not directly deliver.

2) We design a gaze estimation model with head-pose conditioning to improve estimation accuracy and cross-dataset generalization, and construct a structured candidate object set on top of panoptic segmentation that exposes both thing-class instances and stuff-class regions as ready-to-use entities for downstream gaze-target reasoning.

3) We introduce an ambiguity-aware gaze target inference algorithm that adaptively fuses boundary-distance geometry with class-level semantic priors. Beyond the laboratory end-to-end evaluation using Cityscapes road scenes, we further evaluate the modular pipeline on the LBW dataset~\cite{Kasahara2022LBW} after sequence-specific calibration, demonstrating its effectiveness for gaze target identification under naturalistic driving conditions.

The remainder of this paper is organized as follows. Section II reviews related work on driver gaze estimation, semantic understanding of driving scenes, and gaze-scene association. Section III details the proposed framework and the design of each module. Section IV describes the datasets and experimental protocols. Section V reports experimental results. Section VI concludes the paper.

\section{Related Work}

\subsection{Driver Gaze Estimation Techniques}
Gaze estimation is a fundamental component for understanding driver visual attention~\cite{9}. Early studies mainly relied on infrared illumination and dedicated eye trackers, estimating gaze vectors from the geometric relationship between the pupil center and corneal reflection points~\cite{16,17,18}. Although such geometry-based methods can achieve high theoretical accuracy~\cite{19}, the dedicated hardware and cumbersome personalized calibration procedures they require limit their large-scale deployment in real-world driving environments~\cite{1}.

With the development of deep learning, appearance-based gaze estimation has become mainstream, supported by large-scale datasets such as ETH-XGaze~\cite{20}, MPIIFaceGaze~\cite{21}, and Gaze360~\cite{22}. On these benchmarks, CNN-based methods such as Full-Face~\cite{23} and RT-Gene~\cite{24} have substantially advanced appearance-based gaze estimation by regressing gaze direction directly from facial images. It should be noted, however, that these datasets were collected in general, non-driving scenarios, and that most existing methods are trained and evaluated within a single dataset. Whether the resulting models generalize across datasets, and ultimately to real driving, therefore remains to be fully validated.

Furthermore, the local receptive fields of CNNs limit their ability to capture global facial cues, making them less robust to illumination changes, partial occlusions, and large head pose variations~\cite{9}. To address this limitation, recent studies have introduced Vision Transformer architectures~\cite{25} into gaze estimation, enhancing feature modeling along both the spatial and the temporal dimensions. For example, GazeTR~\cite{26} employs self-attention to model long-range dependencies among facial regions; Hybrid ViTNet~\cite{27} adopts a hybrid CNN–Transformer representation to further strengthen appearance-based feature modeling; and FIFA~\cite{28} introduces a fine-grained inter-frame attention mechanism that models pupil displacement across frames for driver video gaze estimation.

Despite the steady progress these methods have brought to gaze estimation accuracy, their generalization under cross-dataset domain shift, and their reliability once transferred to the conditions of real driving, remain insufficiently explored.

\subsection{Semantic Understanding of Driving Scenarios}
Semantic understanding of driving scenes provides the scene-level object information required for driver gaze target identification. Existing road scene parsing methods mainly include object detection, semantic segmentation, instance segmentation, and panoptic segmentation. Object detection methods such as Faster R-CNN~\cite{31} and DETR~\cite{32} can efficiently localize traffic objects with bounding boxes. However, bounding boxes inevitably contain background pixels and cannot accurately describe object contours, which may introduce ambiguity in gaze-target matching. Semantic segmentation methods such as FCN~\cite{33} and DeepLab~\cite{34} provide pixel-wise class labels for the whole scene, but they cannot distinguish different instances within the same class. Instance segmentation methods such as Mask R-CNN~\cite{35} generate independent masks for countable foreground objects, enabling instance-level reasoning, but they offer limited representation for continuous stuff regions such as roads, sky, buildings, and vegetation. In contrast, panoptic segmentation, formally introduced by Kirillov et al.~\cite{36}, unifies instance-level representation for thing classes and semantic-level representation for stuff classes within a single framework, and has since become a mainstream paradigm for urban driving scene understanding on benchmarks such as Cityscapes~\cite{37}.

For driver gaze target identification, however, a pixel-level segmentation map alone is not sufficient. Determining which scene target the driver is looking at requires comparing the gaze location against individual objects, yet a dense segmentation map represents the scene as a continuous field of labeled pixels rather than as separable entities that can be indexed and compared independently. This gap between pixel-level segmentation output and an object-level scene representation remains an obstacle to associating driver gaze with a specific scene target.

\subsection{Gaze–Scene Association for Driver Attention}

Beyond gaze estimation and scene parsing, driver gaze target identification requires associating the estimated gaze with the corresponding target in the forward-view scene. Existing studies on gaze–scene association have evolved from coarse region-level classification to continuous spatial localization, and more recently to attention modeling with scene semantics and class-level priors.

Early studies formulated driver attention estimation as a region-level classification task, in which the driver’s in-cabin field of view was divided into several predefined gaze zones. For example, Fridman et al.~\cite{10} classified driver gaze into six fixed regions, including the instrument panel, rearview mirror, and windshield. Tawari et al.~\cite{11} introduced a distributed multi-camera framework that uses head-pose dynamics for continuous gaze-zone estimation. Zhang et al.~\cite{12} adopted a deep learning strategy by fusing eye, face, and head images within a four-channel convolutional neural network to improve gaze-zone classification accuracy. 

To provide finer-grained driver attention representation than region-level classification, subsequent studies have focused on continuous spatial localization. These methods map gaze direction, head pose, or facial features onto the forward-view road scene to estimate 2D gaze points or attention heatmaps. Recasens et al.~\cite{38} predicted gaze heatmaps in video frames using convolutional networks. Dua et al.~\cite{39} built a driver gaze-to-road-scene mapping dataset and proposed I-DGAZE for gaze point prediction. Hu et al.~\cite{15} developed a calibration-free attention estimation method based on scene images and gaze probability maps. Araluce et al.~\cite{40} combined OpenFace gaze direction with road-scene semantic segmentation through a NARMAX model for gaze focalization in accidental driving scenarios. Although these methods provide higher spatial resolution than region-level classification, their outputs remain limited to spatial representations, rather than explicitly identifying the specific scene target attended to by the driver.

Another line of research is based on the observation that driver attention varies across object categories. Task-relevant objects, such as vehicles, pedestrians, and traffic lights, generally attract more driver attention than background regions, such as the sky or buildings. This tendency is consistent with the findings of Stapel et al.~\cite{41}, who combined eye tracking with automated road-scene perception in a naturalistic on-road study. They found that drivers fixated within 2° of relevant road users much more frequently than irrelevant ones (72.8\% versus 27.8\%), suggesting that gaze behavior reflects the task relevance of surrounding traffic participants. More broadly, Kotseruba and Tsotsos~\cite{3} reviewed evidence showing that driver gaze allocation is shaped by driving tasks and traffic context, supporting the use of task relevance and scene semantics as informative cues for driver attention modeling. Building on this observation, recent studies have incorporated scene semantics, visual saliency, and driving-task priors into driver attention prediction. Palazzi et al.~\cite{13}, Karuppasamy et al.~\cite{42}, and Xu et al.~\cite{14} demonstrated that semantic context and task-relevant information can improve the accuracy of predicted driver attention. In addition, end-to-end models such as EraW-Net~\cite{43} jointly encode driver gaze behavior and road-scene semantics within a unified network architecture.

Despite substantial progress, object-level driver gaze target identification is still underexplored, particularly under a low-cost dual-camera-based setting. Recent object-aware studies have moved closer to this goal by incorporating attended-object cues or scene semantics, but they often focus on attention prediction from road scenes, assume an available gaze point, or do not fully address the ambiguity of gaze-object association in dense traffic scenes. Identifying the specific vehicle, pedestrian, traffic sign, or scene region attended to by the driver requires an integrated framework that combines in-cabin gaze sensing, panoptic scene understanding, gaze-to-scene calibration and mapping, and geometry-semantic adaptive inference. This is the gap addressed in this work.

\section{Research Methodology}

\begin{figure*}[!t]
\centering
\includegraphics[width=\textwidth]{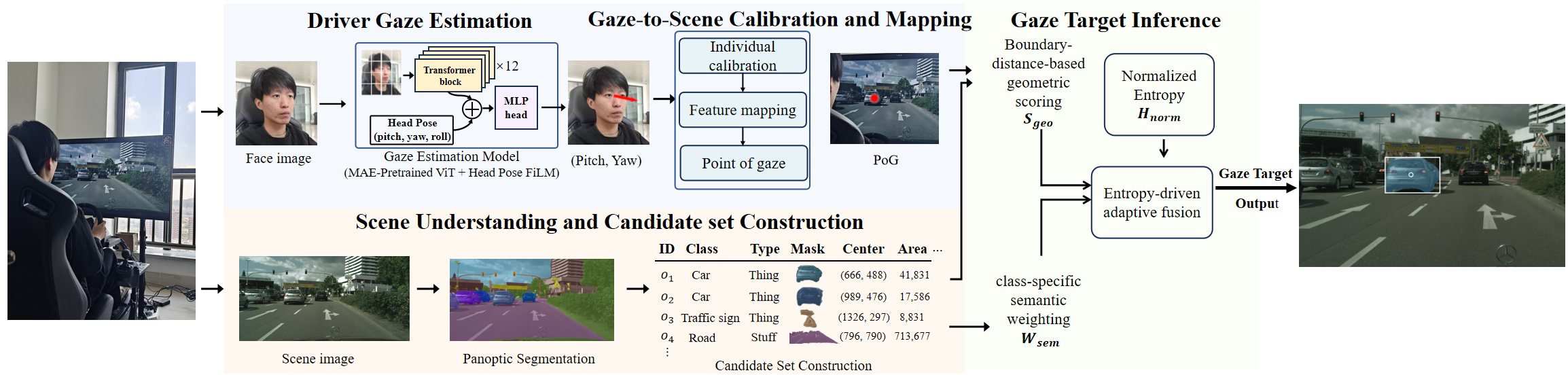}
\caption{Overall framework of the proposed driver gaze target identification method, comprising four cooperating modules: (i) a \textit{gaze estimation module (MFGaze)} that regresses gaze direction from the in-cabin face image with an MAE-pretrained ViT backbone and FiLM-based head-pose modulation; (ii) a \textit{driving-scene understanding module} that applies panoptic segmentation to the forward-view image and reorganizes it into a structured candidate object set; (iii) a \textit{gaze-to-scene calibration and mapping module} that maps gaze features to a 2D point of gaze (PoG) in the scene image; and (iv) a \textit{gaze target inference module} that fuses the boundary-distance geometric score $S_{\mathrm{geo}}$ with class-level semantic weights $W_{\mathrm{sem}}$ through an entropy-driven mechanism governed by the normalized entropy $H_{\mathrm{norm}}$.}
\label{fig:framework}
\end{figure*}

To identify the specific scene target attended to by the driver, this study proposes a gaze target identification framework that integrates high-accuracy gaze estimation with a structured representation of the driving scene. As shown in Fig.~\ref{fig:framework}, the framework takes two parallel visual streams as input. From the in-cabin stream, captured by a driver-facing camera, a stable gaze representation is extracted; from the forward-view stream, captured by a forward-facing camera, a candidate object set with semantic and spatial attributes is constructed. The two outputs are then jointly processed in a unified coordinate space to yield the final gaze target. The framework comprises four cooperating modules: a gaze estimation model that regresses the gaze direction from the face image, a scene understanding module that organizes the road image into a candidate object set, a calibration and mapping module that maps gaze features to 2D scene-image coordinates, and a target inference module that determines the attended target from geometric and semantic evidence. These modules are detailed in the following subsections.

\subsection{Driver Gaze Estimation Model}
The objective of the driver gaze estimation model is to predict the driver's gaze direction from facial images and provide reliable gaze cues for subsequent gaze target identification. Real-world driving scenes often involve illumination changes, partial occlusions, and head pose variations, making methods based on local appearance features prone to unstable feature representations. To improve robustness in complex driving scenes, this study adopts a gaze estimation model with a ViT backbone that incorporates head pose information to adaptively modulate visual features.

\begin{figure*}[!t]
    \centering
    \includegraphics[width=0.8\linewidth]{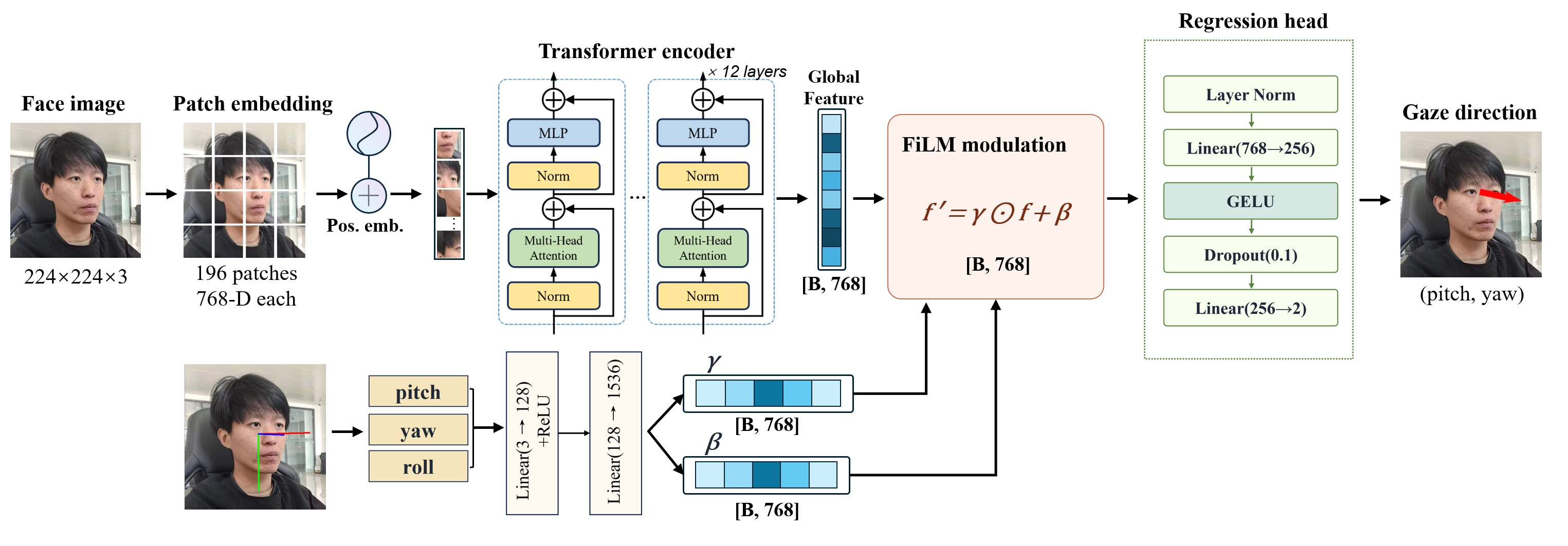}
     \caption{Architecture of the proposed driver gaze estimation model. The normalized $224\times224$ driver face image is divided into $14\times14$ grid of patches and processed by an MAE-pretrained ViT-B/16 backbone to extract a 768-dimensional global visual feature. Head pose information is fed into a two-layer MLP to generate the FiLM scale and shift parameters, $\gamma$ and $\beta$. The visual feature is adaptively modulated according to the head pose condition and then passed through a lightweight regression head to predict the gaze direction.}
    \label{fig:model}
\end{figure*}

As shown in Fig.~\ref{fig:model}, the model comprises a facial feature extraction branch, a head-pose modulation branch, and a gaze regression head. The normalized facial image is processed by the ViT backbone to extract global visual features. Through self-supervised pretraining and the self-attention mechanism, the backbone captures long-range dependencies across different facial regions, including the eyes, facial contours, and overall facial structure. This design aims to improve the robustness of gaze representations under illumination changes and partial occlusions. Concurrently, head pose information is introduced as a geometric prior, and a multilayer perceptron is employed to generate the scale and shift parameters of the FiLM modulation. The visual features are then adaptively modulated according to the estimated head pose, thereby enabling the model to better capture the interaction between head orientation and gaze direction. The model is optimized using an L1 loss between the predicted and ground-truth gaze angles. The predicted gaze direction serves as the gaze cue for subsequent gaze-to-scene mapping and gaze target identification.

\subsection{Driving Scene Understanding and Candidate Set Construction}
To achieve driver gaze target identification, the system must not only obtain the driver's gaze direction but also extract and analyze potential gaze targets in the road scene ahead. The task of this module is to extract a structured scene representation from the input road image and provide a candidate object set for the subsequent gaze target inference.

Considering that road scenes contain both discrete objects with distinct boundaries (such as vehicles, pedestrians, and traffic signs) and amorphous background regions (such as roads, sky, and vegetation), this paper adopts a panoptic segmentation framework to uniformly process both types of objects. Specifically, this paper uses Mask2Former with ResNet-50 backbone network as the panoptic segmentation model, and loads the official pre-trained weights from the Cityscapes panoptic segmentation dataset. The model outputs a panoptic segmentation map for the input road scene image. Each "thing" class instance (car, person, etc.) is assigned a unique instance ID, while "stuff" class regions (road, sky, etc.) are represented by their semantic class. After processing by the scene understanding module, the original road image is converted into a structured set of candidate objects $O = \{o_1, o_2, \ldots, o_N\}$.
Each candidate object is represented as:
\begin{equation}
{o_i} = \{ i{d_i},{c_i},{m_i},{p_i},{s_i}\}
\label{eq:candidate_set}
\end{equation}
where \(id_i\) denotes the unique identifier, \(c_i\) denotes the semantic class, \(m_i\)denotes the segmentation mask, \(p_i\) denotes the object center position, and \(s_i\) denotes the object scale measured as the pixel area of the segmentation mask. This structured representation encodes both spatial information (position and scale) and semantic information (class), providing comprehensive input for subsequent gaze target inference.
\subsection{Gaze-to-Scene Calibration and Mapping}
After driver gaze estimation and scene understanding are completed, the two outputs must be unified into the same coordinate space for association inference. The gaze estimation model outputs a 3D gaze direction in the Camera Coordinate System (CCS), whereas candidate targets are represented in the two-dimensional scene image. A learning-based, individualized calibration method therefore maps gaze features to the driver's 2D Point of Gaze (PoG) coordinates in this image. In the simulator experiment, these coordinates correspond to the displayed road scene on the screen; in the LBW evaluation, they correspond to the forward-view image. The mapping is fitted for each participant in the simulator experiment and for each recording sequence in the LBW evaluation, as described in Section IV. To characterize the factors influencing the PoG, an 8-dimensional input feature vector is constructed:
\begin{equation}
\mathbf{x} =
[\theta_p, \theta_y, h_x, h_y, h_z, \phi_p, \phi_y, \phi_r]^T
\in \mathbb{R}^{8}
\label{eq:2}
\end{equation}
where \(\theta_p\) and \(\theta_y\) are the pitch and yaw angles output by the gaze estimation model; (\(h_x, h_y, h_z\)) are the 3D coordinates of the driver’s head center in the CCS obtained during geometric normalization; \(\phi_p\), \(\phi_y\), and \(\phi_r\) are the pitch, yaw, and roll angles of the head pose. Head position helps compensate for the gaze-origin offset caused by differences in driver seating posture, and head pose further captures the geometric influence of head orientation on the PoG.

To fit the nonlinear mapping from input features to 2D scene-image coordinates, this study employs a second-order polynomial ridge regression model. First, the input features are z-score standardized to remove scale differences among features:
\begin{equation}
\widetilde {\bf{x}} = ({\bf{x}} - \mu )/\sigma 
\label{eq:3}
\end{equation}

A second-order polynomial feature expansion is then applied to obtain high-dimensional features consisting of the original features, squared terms, and pairwise interaction terms:
\begin{equation}
\phi (\widetilde {\bf{x}}) = {[{\tilde x_1},\; \ldots ,\;{\tilde x_8},\;\tilde x_1^2,\; \ldots ,\;\tilde x_8^2,\;{\tilde x_1}{\tilde x_2},\; \ldots ,\;{\tilde x_7}{\tilde x_8}]^T} \in \mathbb{R}^{44}
\label{eq:4}
\end{equation}

Because the horizontal and vertical scene-image coordinates are affected differently by the input features, separate ridge regression models are trained for the normalized horizontal coordinate and the normalized vertical coordinate. Taking the horizontal coordinate as an example, the predicted value is obtained by linearly weighting the polynomial features:
\begin{equation}
\hat u' = {\bf{w}}_u^T\phi (\widetilde {\bf{x}}) + {b_u}
\label{eq:5}
\end{equation}

The model is optimized by minimizing the following loss function:

\begin{equation}
L = \frac{1}{M}\sum\limits_{k = 1}^M {{\omega _k}} {({u'_k} - {\hat u'_k})^2} + \alpha \| \mathbf{w}_u \|_2^2
\label{eq:6}
\end{equation}
where $M$ is the number of calibration samples, and $\alpha$ is the regularization coefficient used to constrain the norm of the weight vector and prevent overfitting in the high-dimensional polynomial feature space. In the simulator experiment, peripheral screen regions naturally exhibit higher gaze estimation and mapping errors than the central region, owing to greater head rotation and more pronounced perspective distortion in eye images. 

To account for this issue, this study introduces a sample weight $\omega_k$ into the training loss. The sample weight $\omega_k$ is assigned according to:
\begin{equation}
\omega_k = 1 + \eta\, r
\label{eq:7}
\end{equation}
where $r$ is the normalized distance from the target point to the screen center, and $\eta$ is a positive coefficient that assigns higher weights to calibration samples near the screen edges; its value in the simulator experiment is reported in Section V-B.

\subsection{Gaze Target Inference}
Following the preceding subsections, both the driver's scene-image PoG $P$ and the candidate object set $O$ in the current road scene have been obtained. The task of this section is to combine these two types of information and determine the object that the driver is most likely gazing at from the candidate object set.

Because panoptic segmentation performs a dense partition of the road scene, any location in the scene image necessarily falls within the mask of a certain object or region, and there are no gaps between adjacent masks. This property makes direct mask-hit judgment overly sensitive: a slight deviation of the gaze point from the true object contour may cause the result to switch to an adjacent candidate. In addition, drivers exhibit systematic attention preferences toward different object classes. For example, pedestrians and vehicles usually receive much more attention than static backgrounds such as the sky or buildings. Relying solely on geometric information ignores this regularity. Based on these considerations, this section introduces both spatial-geometric metrics and class-level semantic priors to balance spatial accuracy and semantic plausibility.

\subsubsection{\textbf{Boundary-Distance-Based Geometric Matching Score}}
The goal of geometric matching is to quantify the spatial proximity between each candidate object and $P$. This study uses the shortest distance from $P$ to the object mask boundary as the basic metric. When $P$ lies inside the object, the distance is set to 0; when $P$ lies outside the object, the distance is defined as the Euclidean distance from $P$ to the nearest point on the object contour. Compared to the centroid distance, the boundary distance treats objects of different sizes more fairly. It avoids the underestimation of large objects whose centroids are far from $P$ and reduces errors for elongated objects whose centroids may lie outside the visually attended region.

The boundary distance is converted into a geometric matching score using an exponential decay function:
\begin{equation}
S_{\mathrm{geo}}(i) = \exp\left(-\frac{d_{\mathrm{edge}}(i)}{\tau}\right)
\label{eq:geo_score}
\end{equation}
where $\tau$ is the decay scale parameter, controlling the rate at which the score decreases with distance; its appropriate value should be of the same order as the total error of the line-of-sight estimation and calibration modules in this framework. The specific value of $\tau$ is determined experimentally within a reasonable range.
In real-world driving scenes, panoptic segmentation may generate a large number of candidate objects. Directly normalizing the geometric matching scores over all candidates would allow distant objects to occupy a non-negligible portion of the probability mass, since they are numerous even though their individual scores are small. Therefore, this study retains only candidate objects with a boundary distance no greater than 5$\tau$ for subsequent decision-making, forming an effective candidate set $O_{\mathrm{eff}}$. Candidates beyond $5\tau$ have geometric scores below $e^{-5}\approx0.0067$, motivating this practical cutoff to limit the contribution of distant candidates. The geometric matching scores of the effective candidate set are then normalized to obtain the geometric probability distribution.

\subsubsection{\textbf{Class-Specific Semantic Weights}}
Drivers exhibit significant differences in attention allocation across different object classes. Multiple eye-tracking studies based on real-world driving have shown that task-relevant objects with potential interactive significance, such as pedestrians and vehicles, receive longer fixation durations and higher fixation frequencies from drivers. This pattern suggests that when there is ambiguity regarding the gaze landing point, the relevance of the category itself can provide valuable supplementary information for decision-making.

Based on this observation, a semantic weight $W_{\mathrm{sem}}(c)$ is assigned to each category of candidate objects to reflect its relative importance in the driving task. Let the category of candidate object $o_i$ be $c_i$, and its semantic prior score is defined as
\begin{equation}
S_{\mathrm{sem}}(i) = W_{\mathrm{sem}}(c_i)
\label{eq:semantic_score}
\end{equation}

For the semantic classes considered in this study, this paper assigns corresponding weights based on their roles in driving tasks, as illustrated in Table~\ref{tab:semantic_weights}.

\begin{table}[!t]
\centering
\caption{Semantic Weights of Candidate Object Categories}
\label{tab:semantic_weights}
\renewcommand{\arraystretch}{0.9}
\begin{tabular}{llc}
\toprule
\textbf{Group} & \textbf{class} & \textbf{Weight} \\
\midrule
\multirow{5}{*}{Task-relevant objects}
& person        & 1.5 \\
& car           & 1.4 \\
& bicycle       & 1.3 \\
& traffic sign  & 1.3 \\
& traffic light & 1.3 \\
\midrule
\multirow{5}{*}{Background regions}
& road       & 1.0 \\
& sidewalk   & 0.9 \\
& vegetation & 0.8 \\
& building   & 0.7 \\
& sky        & 0.5 \\
\bottomrule
\end{tabular}
\end{table}

The weight values are heuristically defined based on prior knowledge of driver attention behavior. The sensitivity analysis in Section V further verifies the robustness of final identification accuracy to weight perturbations, confirming that the weight setting is guided by empirical knowledge rather than the precise selection of specific numerical values.

\subsubsection{\textbf{Entropy-driven adaptive fusion}}

The geometric and semantic terms provide two types of information: instantaneous spatial evidence and stable class-level priors. Their relative importance should be dynamically adjusted according to the clarity of geometric matching in the current frame. When geometric information is sufficient to distinguish the gazed object, semantic priors should only play a supporting role; when geometric information is ambiguous, semantic priors should intervene more strongly in decision-making. This study uses normalized entropy to characterize the certainty of geometric matching and dynamically adjusts the fusion coefficient $\lambda$, which controls the influence strength of the semantic prior. 

\begin{figure}[!t]
\centering
\includegraphics[width=\linewidth]{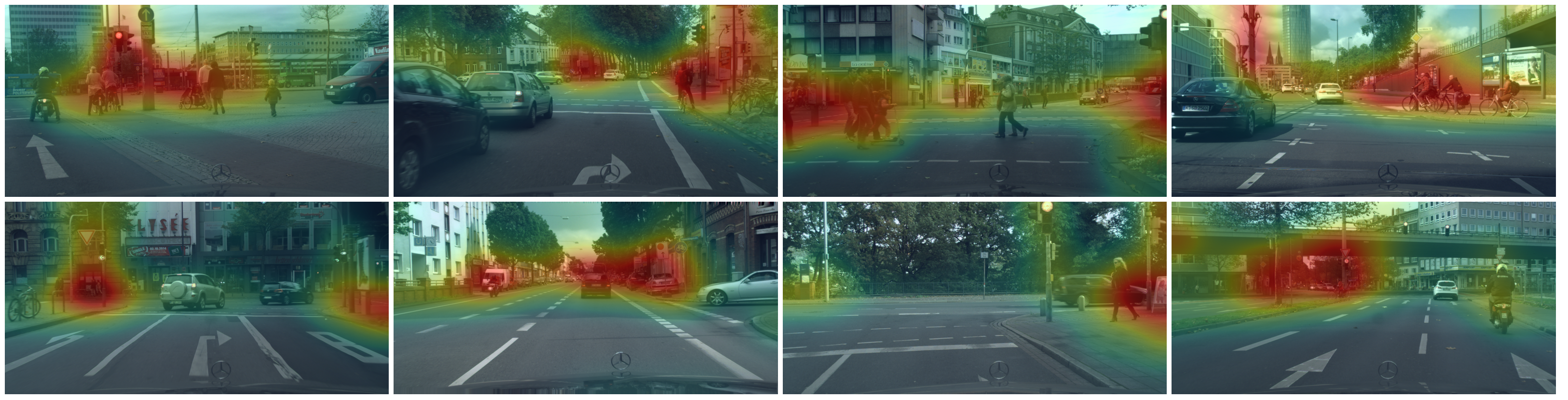}
\caption{Spatial visualization of the normalized entropy $H_{\mathrm{norm}}$ over representative road scenes. Within each scene, regions with sparsely distributed objects exhibit low $H_{\mathrm{norm}}$ (unambiguous geometric landing), whereas regions with densely clustered objects exhibit high $H_{\mathrm{norm}}$ (ambiguous geometric landing).}
\label{fig:Hnorm}
\end{figure}

For the geometric probability distribution $\tilde{S}_{\mathrm{geo}}(i)$ over the set of valid candidates, the normalized entropy is defined as:
\begin{equation}
H_{\mathrm{norm}}
=
-\frac{1}{\log N_{\mathrm{eff}}}
\sum_{i=1}^{N_{\mathrm{eff}}}
\tilde{S}_{\mathrm{geo}}(i)
\log \tilde{S}_{\mathrm{geo}}(i)
\label{eq:normalized_entropy}
\end{equation}
where $N_{\mathrm{eff}}$ is the number of valid candidate objects, and $\log N_{\mathrm{eff}}$ is the maximum entropy under a uniform distribution, which normalizes the entropy value to the interval $[0,1]$. When $N_{\mathrm{eff}}=1$, $H_{\mathrm{norm}}$ is defined as 0. Through normalization, the ambiguity measure maintains consistent physical meaning across scenes with different numbers of candidate objects. As shown in Fig.~\ref{fig:Hnorm}, within a single scene, sparsely distributed objects yield a low $H_{\mathrm{norm}}$ and a clear geometric match, whereas densely clustered objects yield a high $H_{\mathrm{norm}}$, where the semantic prior should intervene more strongly.

The fusion coefficient $\lambda$ is calculated through a linear mapping from the normalized entropy:
\begin{equation}
\lambda =
\lambda_{\min}
+
(\lambda_{\max} - \lambda_{\min}) H_{\mathrm{norm}}
\label{eq:fusion_coefficient}
\end{equation}
where $\lambda_{\min}$ and $\lambda_{\max}$ denote the lower and upper bounds of the semantic prior influence, respectively. The lower bound guarantees a minimum influence of the semantic prior even when geometric evidence is clear, so that the decision is not dominated by negligible geometric differences when candidate classes of different semantic weights occupy similar areas. The upper bound limits the maximum influence of the semantic prior in highly complex scenes, preventing the decision from becoming detached from spatial evidence. In this study, $\lambda_{\min}$ is set to $0.2$ and $\lambda_{\max}$ is set to $1.0$. This range covers the reasonable interval for semantic-prior influence, and its validity is further verified by the sensitivity analysis in Section~\ref{sec:results}.

The final decision score is calculated from the geometric probability and semantic weight:
\begin{equation}
S_{\mathrm{final}}(i)
=
\tilde{S}_{\mathrm{geo}}(i)
\cdot
W_{\mathrm{sem}}(c_i)^{\lambda}
\label{eq:final_score}
\end{equation}

\section{Datasets and Experimental Protocols}

This section describes the datasets and experimental protocols used to evaluate the proposed framework at different stages. Section IV-A introduces the three public datasets used to train and evaluate the gaze estimation module. Sections IV-B and IV-C describe the self-collected participant data used for gaze-to-screen calibration and gaze target identification, respectively. Section IV-D introduces the public LBW dataset and the protocol used to evaluate the proposed pipeline under naturalistic driving conditions.

\subsection{Public Gaze Estimation Datasets}

This study trained gaze estimation models on three public datasets: ETH-XGaze, MPIIFaceGaze, and Gaze360. These datasets complement each other in terms of head pose ranges, illumination, and scene diversity. ETH-XGaze covers a wide range of head poses, including the large head turns that occur in driving scenes. MPIIFaceGaze provides face images captured under diverse natural lighting conditions, reflecting the varying illumination encountered in real-world driving. Gaze360 spans both indoor and outdoor settings, and its scene diversity supports generalization to uncontrolled environments. ETH-XGaze is divided into training, validation, and test sets in an 8:1:1 ratio; MPIIFaceGaze follows the Leave-One-Person-Out (LOPO) protocol; and Gaze360 adopts its official training and test split. A geometric normalization process is applied to all datasets to obtain facial images of $224\times224$ pixels.

\subsection{Data Collection for Gaze-to-Screen Calibration}

\begin{figure}[!t]
\centering
\subfloat[Experimental setup]{
\includegraphics[width=0.46\linewidth]{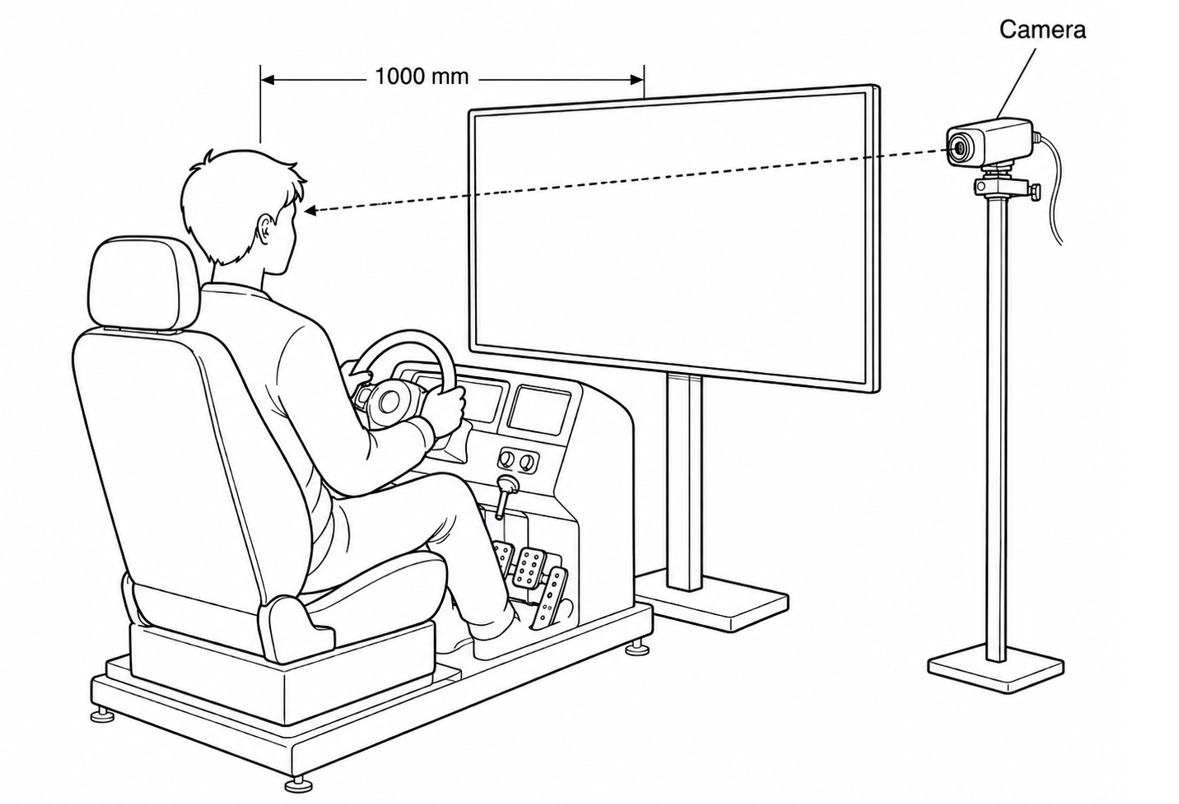}
\label{fig:moniqi}
}
\hfill
\subfloat[Calibration interface]{
\includegraphics[width=0.46\linewidth]{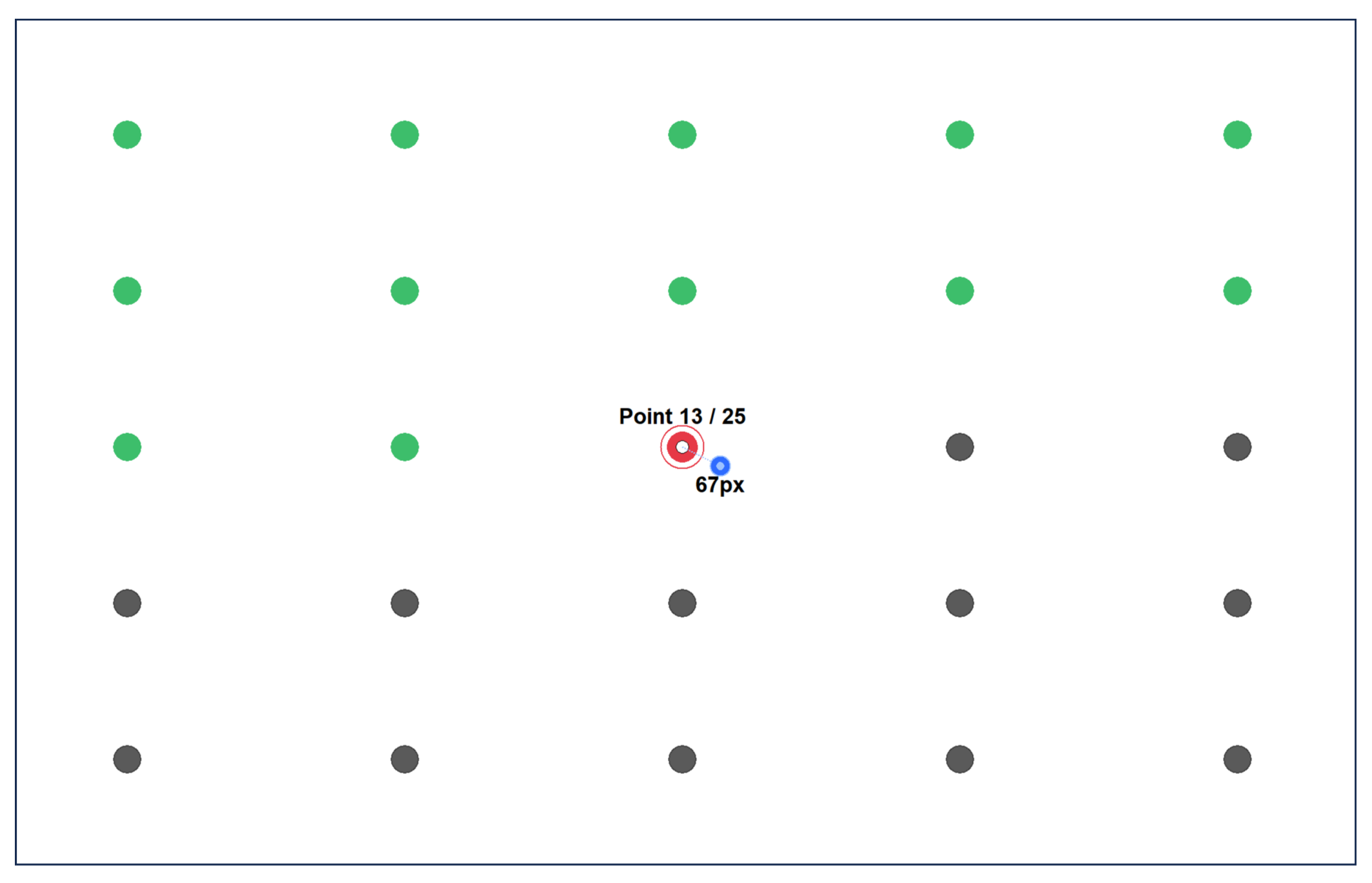}
\label{fig:biaoding}
}
\caption{Overview of the experimental environment and calibration procedure. (a) Schematic of the driving simulator and camera setup, where the participant's face is approximately 1000 mm from the screen. (b) The calibration interface displayed on the simulator screen. The red dot indicates the current calibration target, green dots indicate completed targets, and the blue dot represents the real-time predicted point of gaze based on the completed targets.}
\label{fig:calibration}
\end{figure}

The calibration data were collected on a driving simulator platform. The experiment used a 27-inch simulator display, with the face-image capture camera operating at a resolution of $1280\times720$ pixels. The camera intrinsic parameters were pre-determined using checkerboard calibration. Calibration and the subsequent gaze target identification test were performed consecutively within the same session, ensuring that the participant's seating posture and head position remained essentially consistent across both phases. Schematic illustrations of the experimental setup and calibration interface are presented in Fig.~\ref{fig:calibration}.

A $5\times5$ grid of 25 calibration targets was displayed on the screen in a serpentine scanning sequence. For each target, face images were captured over 2.5\,s, from which 15 valid frames were retained. Each captured frame was processed by face detection, facial landmark localization, and geometric normalization, and was retained as a valid sample only when all three steps succeeded. After geometric normalization, a $224\times224$-pixel face image was obtained, together with the corresponding 3D head position and head pose. The face image was then fed into the gaze estimation model to obtain the predicted gaze direction (pitch and yaw). The gaze direction, head position, and head pose were concatenated in the order defined in Section III-C to construct an 8-dimensional input feature vector, which was paired with the normalized screen coordinates of the corresponding target point to form a single calibration sample.

\subsection{Data Collection for Gaze Target Identification}
An end-to-end test dataset was collected using the driving simulator to evaluate the complete pipeline from synchronized driver-face and road-scene images to the identified gaze target. The forward-view road-scene images were selected from the Cityscapes dataset, while the corresponding driver-face images and ground-truth target labels were collected during the experiment. The selected scenes covered diverse object categories and candidate densities, including both open road sections with sparse targets and complex scenes containing densely distributed objects.

After panoptic segmentation, each selected road-scene image was displayed on the simulator screen. Candidate objects were highlighted sequentially in a predefined order, and the participant was instructed to fixate on each highlighted candidate during its display period. Five consecutive driver-face images were captured for each candidate. Each valid face image was paired with the ID of the currently highlighted object, forming a ground-truth sample for end-to-end gaze target identification.

\begin{figure}[!t]
    \centering
    
    \begin{minipage}[c]{0.48\linewidth}
        \centering
        \subfloat[Panoptic Segmentation]{
            \includegraphics[width=\linewidth]{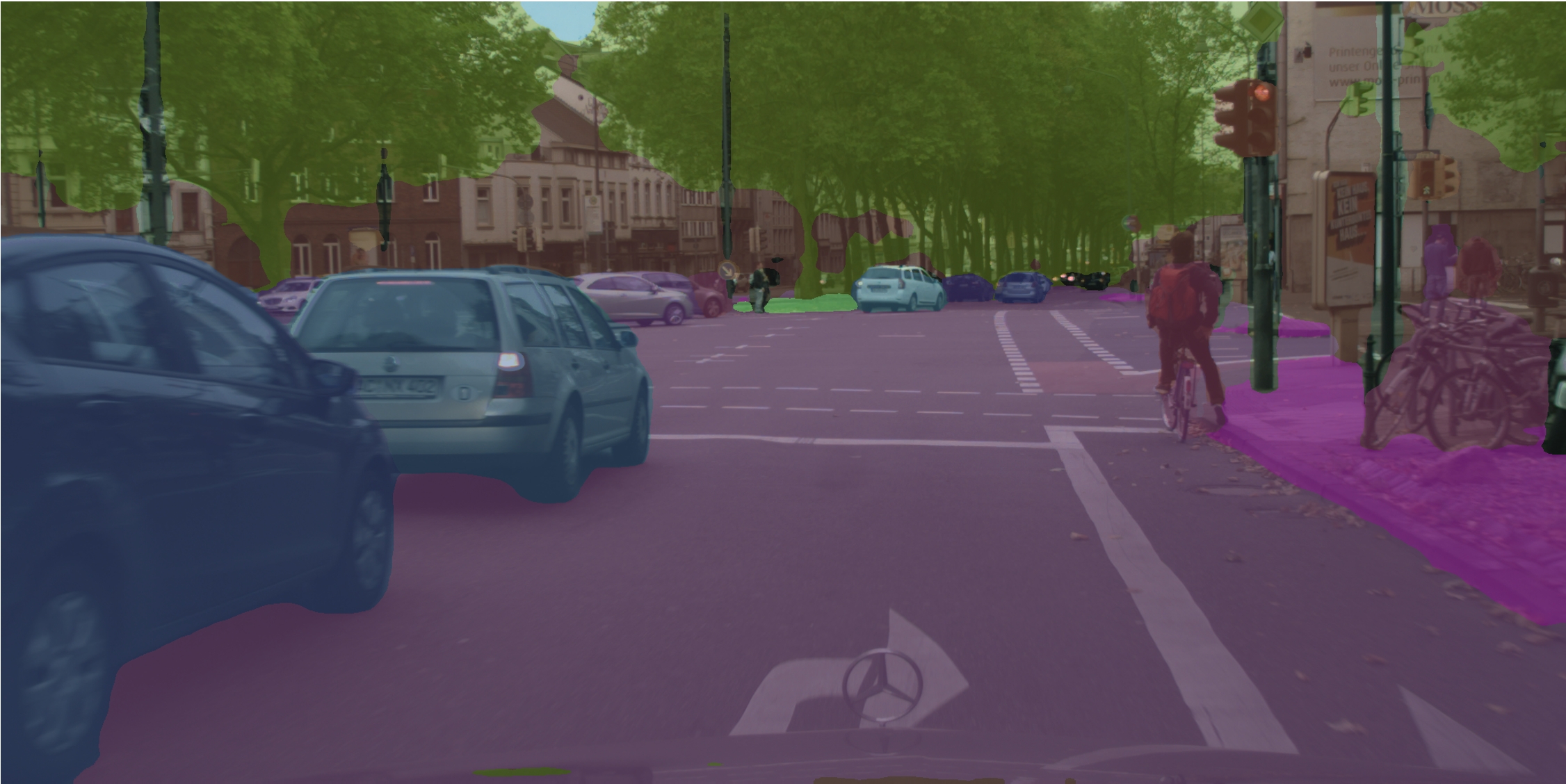}
            \label{fig:fenge}
        }
        
        \vspace{0.2cm} 

        \subfloat[Target Identification]{
            \includegraphics[width=\linewidth]{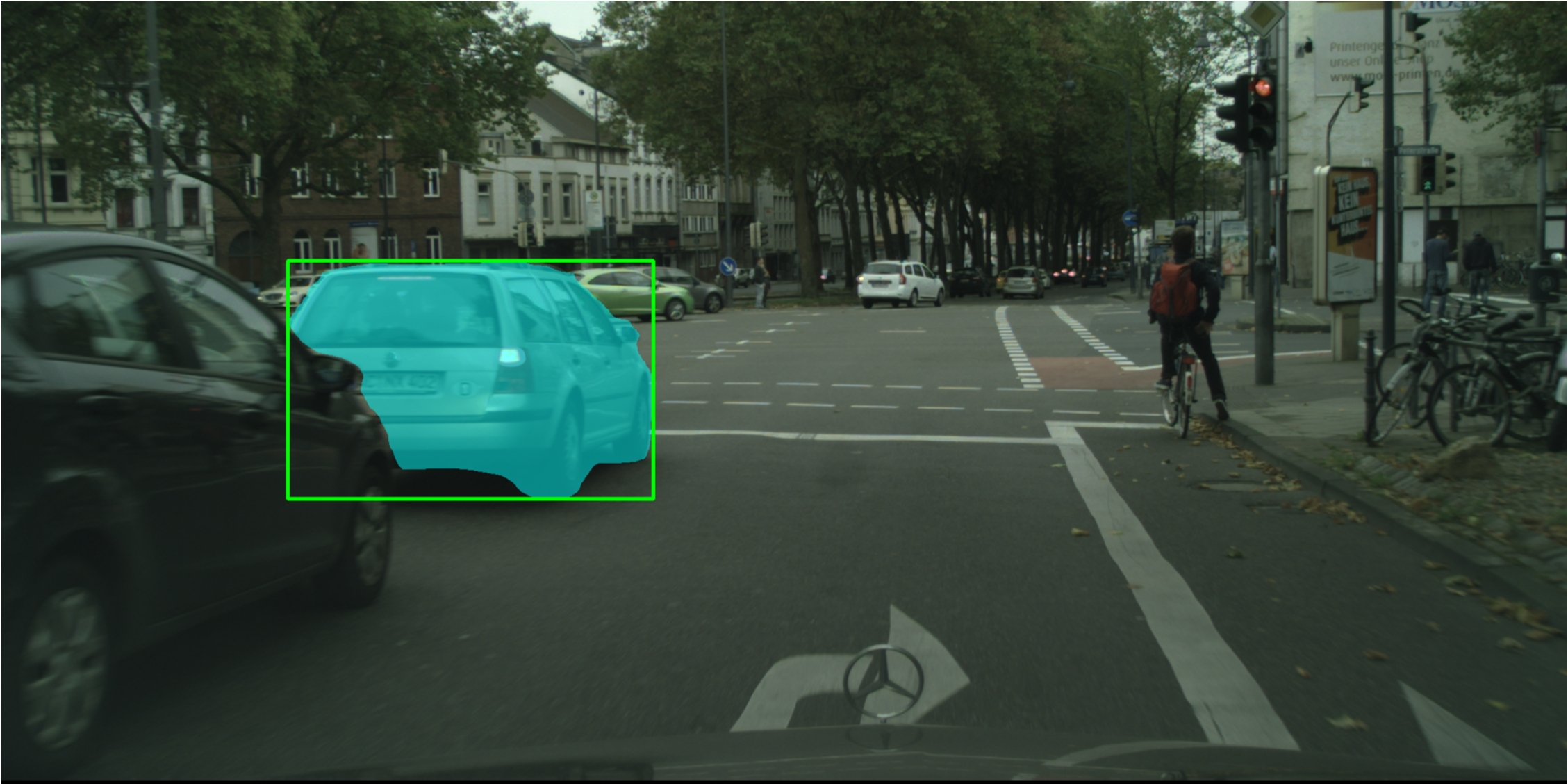}
            \label{fig:dange}
        }
    \end{minipage}
    \hfill                                     
    \begin{minipage}[c]{0.48\linewidth}
        \centering
        \subfloat[Cropped targets and corresponding driver faces]{
            \includegraphics[width=\linewidth]{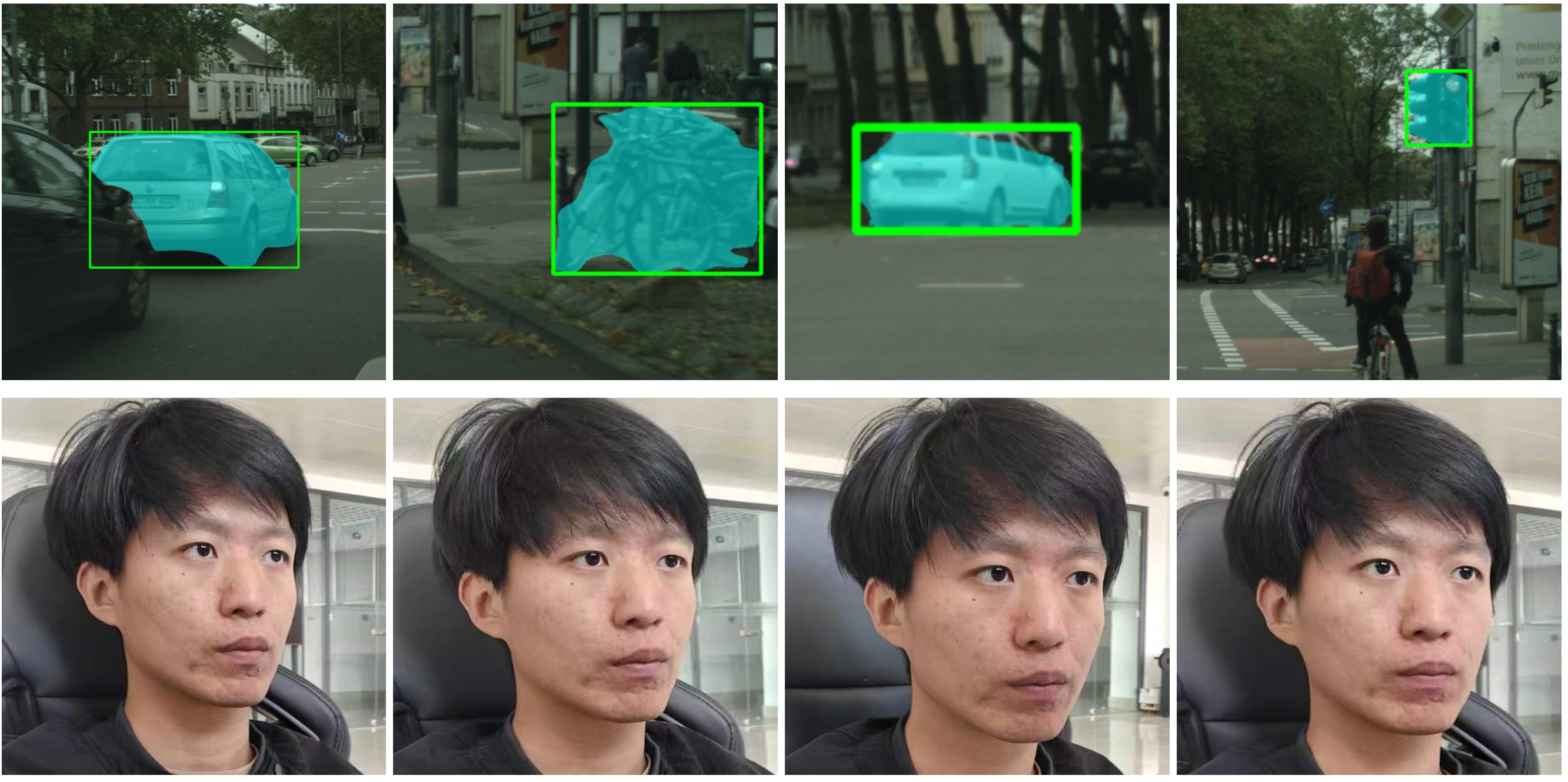}
            \label{fig:duizhao}
        }
    \end{minipage}
    
    \caption{One end-to-end test sample. (a) Panoptic segmentation of a representative Cityscapes scene, from which the structured candidate set is extracted. (b) The candidate target highlighted in the full road scene for participant fixation. (c) Cropped close-ups of several such fixation targets paired with the corresponding driver faces, forming the ground-truth pairs used for end-to-end evaluation.}
    \label{fig:three_images_layout}
\end{figure}

Fig.~\ref{fig:three_images_layout} illustrates a representative end-to-end test sample. Fig.~\ref{fig:fenge} presents the panoptic segmentation result of a Cityscapes scene, from which the structured candidate set was extracted. Fig.~\ref{fig:dange} shows the candidate object highlighted as the fixation target, whereas Fig.~\ref{fig:duizhao} presents examples of highlighted targets paired with the corresponding synchronized driver-face images, forming the ground-truth samples used for end-to-end evaluation.

Given their different roles in driving scenes, the samples were organized by semantic class and further divided into task-relevant objects and background regions. Task-relevant objects included car, person, traffic sign, traffic light, and bicycle, whereas background regions included road, sidewalk, building, vegetation, and sky. Both class-wise and group-level results were reported to evaluate the target inference methods across different semantic classes and target groups.

\subsection{LBW Dataset and Evaluation Protocol}

The Look Both Ways (LBW) dataset~\cite{Kasahara2022LBW} is a naturalistic on-road driving dataset that provides synchronized driver-face and forward-view recordings, together with gaze points measured using wearable eye-tracking glasses and registered to the forward-view images. The data used in this study comprise 75 recording sequences from 28 participants and 123,297 synchronized image pairs, with a forward-view resolution of $942\times489$ pixels. Among the publicly available datasets considered, LBW most closely matches the input requirements of the proposed pipeline and was therefore used for evaluation under naturalistic driving conditions. Nevertheless, the relatively low image resolution makes small and distant traffic targets difficult to resolve. In addition, LBW provides the driver's gaze point on each forward-view image rather than the identity of the corresponding gaze target; panoptic segmentation is therefore used to associate each gaze point with its corresponding scene target.

For each sequence, 30\% of the frames were uniformly sampled to fit a participant- and sequence-specific geometric mapping from gaze features to 2D scene coordinates; their gaze points served only as calibration targets. The remaining non-overlapping 70\% formed the evaluation subset; calibration frames were excluded from the reported results, and the target-inference parameters remained fixed. This calibration compensates for individual gaze characteristics and recording-specific driver-camera geometry. Thus, the protocol assesses performance after sequence-specific calibration rather than calibration-free generalization to unseen participants.

For each evaluation frame, the gaze estimation module predicted the gaze direction, which was mapped onto the forward-view image and associated with the candidate targets generated by the scene understanding module. Because LBW provides point-level gaze annotations rather than object-level gaze-target labels, the panoptic segment containing the eye-tracker gaze point was used as the reference target, thereby converting each gaze point into an object- or region-level target label. The analysis focused on the road class and five task-relevant object classes: car, person, bicycle, traffic sign, and traffic light. Frames associated with other scene categories, such as sidewalk, building, vegetation, and sky, were not included in this task-focused analysis.
\section{Experimental Results and Analysis}\label{sec:results}

\subsection{Evaluation of the Gaze Estimation Model}

This section presents a three-tiered evaluation of the gaze estimation model proposed in Section III. First, ablation experiments are conducted to quantify the independent contribution of each design component to model's accuracy. Second, a comparison is made between the proposed model and state-of-the-art methods under identical conditions across three public datasets in order to validate its competitiveness relative to existing work. Finally, the model's generalization ability under domain shift conditions is evaluated through cross-dataset validation.

\subsubsection{\textbf{Ablation Studies}}

To validate the contribution of each core component in the gaze estimation model, an ablation experiment was conducted on the ETH-XGaze dataset. All configurations shared the same MAE-pretrained ViT backbone, training hyperparameters, and L1 loss function, and differed only in the method of head pose fusion. The experiment involved a comparison of three configurations: (1) a Baseline that does not incorporate head pose information; (2) a scheme using Concat fusion; and (3) the FiLM-modulated approach ultimately adopted in this study. In addition to the in-domain test set (ETH-XGaze), this paper also reports test results on the full MPIIFaceGaze dataset to validate the contribution of each component to cross-dataset generalization. The results are shown in Table~\ref{tab:gaze_ablation}.

\begin{table}[!t]
\centering
\caption{Ablation Study Results of the Gaze Estimation Model on the ETH-XGaze Test Set and MPIIFaceGaze Cross-Dataset Evaluation}
\label{tab:gaze_ablation}
\renewcommand{\arraystretch}{1.15}
\footnotesize
\setlength{\tabcolsep}{3pt}
\begin{tabular}{lcc}
\toprule
\textbf{Model Configuration} & \textbf{ETH-XGaze Test} & \textbf{MPIIFaceGaze Cross} \\
\midrule
Baseline & 4.62$^\circ$ & 9.10$^\circ$ \\
Baseline + Concat & 4.31$^\circ$ & 7.31$^\circ$ \\
\textbf{Ours (MFGaze)} & \textbf{4.23$^\circ$} & \textbf{6.93$^\circ$} \\
\bottomrule
\end{tabular}
\end{table}

\begin{table}[!t]
\centering
\caption{Comparison of Mean Angular Error (MAE, $^\circ$) with State-of-the-Art Methods on Public Datasets}
\label{tab:sota_gaze_comparison}
\renewcommand{\arraystretch}{1.15}
\setlength{\tabcolsep}{6pt}
\footnotesize
\begin{tabular}{lccc}
\toprule
\textbf{Model} & \textbf{MPIIFaceGaze} & \textbf{ETH-XGaze} & \textbf{Gaze360} \\
\midrule
ResNet-18 & 5.13$^\circ$ & 4.94$^\circ$ & 13.72$^\circ$ \\
ResNet-50 & 5.19$^\circ$ & 4.83$^\circ$ & 12.11$^\circ$ \\
L2CS-Net  & 4.12$^\circ$ & 5.57$^\circ$ & 16.33$^\circ$ \\
FullFace  & 5.15$^\circ$ & 4.73$^\circ$ & 14.99$^\circ$ \\
Gaze360   & \textit{N/A}& \textit{N/A}& 11.04$^\circ$ \\
GazeTR    & 4.28$^\circ$ & 4.67$^\circ$ & 11.68$^\circ$ \\
\midrule
\textbf{Ours (MFGaze)}      & \textbf{4.06$^\circ$} & \textbf{4.23$^\circ$} & \textbf{10.52$^\circ$} \\
\bottomrule
\end{tabular}
\end{table}

\begin{table*}[!t]
\centering
\caption{Cross-Dataset Evaluation Results of Mean Angular Error (MAE, $^\circ$)}
\label{tab:cross_dataset_results}
\renewcommand{\arraystretch}{1.2}
\setlength{\tabcolsep}{7pt}
\footnotesize
\begin{tabular}{lcccccc}
\toprule
\textbf{Models/Test} 
& \makecell{\textbf{ETH-XGaze}$\rightarrow$\\\textbf{MPIIFaceGaze}} 
& \makecell{\textbf{ETH-XGaze}$\rightarrow$\\\textbf{Gaze360}} 
& \makecell{\textbf{Gaze360}$\rightarrow$\\\textbf{ETH-XGaze}} 
& \makecell{\textbf{Gaze360}$\rightarrow$\\\textbf{MPIIFaceGaze}} 
& \makecell{\textbf{MPIIFaceGaze}\\$\rightarrow$\textbf{ETH-XGaze}} 
& \makecell{\textbf{MPIIFaceGaze}\\$\rightarrow$\textbf{Gaze360}} \\
\midrule
ResNet-18 & 7.56$^\circ$ & 18.68$^\circ$ & 22.95$^\circ$ & 11.15$^\circ$ & \textbf{23.68$^\circ$} & 29.70$^\circ$ \\
ResNet-50 & 7.48$^\circ$ & 20.61$^\circ$ & 18.59$^\circ$ & 11.90$^\circ$ & 23.72$^\circ$ & 27.96$^\circ$ \\
L2CS-Net  & 7.33$^\circ$ & 20.91$^\circ$ & 16.70$^\circ$ & 11.36$^\circ$ & 24.30$^\circ$ & 30.86$^\circ$ \\
GazeTR    & 7.42$^\circ$ & 21.79$^\circ$ & 17.31$^\circ$ & 10.11$^\circ$ & 26.28$^\circ$ & 29.97$^\circ$ \\
FullFace  & 7.33$^\circ$ & 21.50$^\circ$ & 15.56$^\circ$ & 10.42$^\circ$ & 25.32$^\circ$ & 29.24$^\circ$ \\
\textbf{Ours (MFGaze)} & \textbf{6.93$^\circ$} & \textbf{15.64$^\circ$} & \textbf{12.42$^\circ$} & \textbf{7.53$^\circ$} & 23.85$^\circ$ & \textbf{24.15$^\circ$} \\
\bottomrule
\end{tabular}
\end{table*}

In Table~\ref{tab:gaze_ablation}, the incorporation of head pose information consistently improved gaze estimation accuracy over the Baseline model. On the ETH-XGaze test set, Baseline + Concat and MFGaze reduced the mean angular error from 4.62$^\circ$ to 4.31$^\circ$ and 4.23$^\circ$, corresponding to improvements of 0.31$^\circ$ and 0.39$^\circ$, respectively. These results suggest that head pose serves as an effective geometric prior, providing additional constraints for gaze direction prediction and alleviating the ambiguity caused by head rotations when only facial appearance features are used.

Beyond this overall gain, the advantage of FiLM modulation over simple concatenation was most evident under cross-dataset evaluation. On the ETH-XGaze test set, MFGaze reduced the error only marginally over Baseline + Concat, from 4.31$^\circ$ to 4.23$^\circ$ (0.08$^\circ$), whereas on the MPIIFaceGaze cross-dataset test it reduced the error from 7.31$^\circ$ to 6.93$^\circ$ (0.38$^\circ$). This contrast indicates that simple concatenation can incorporate head pose information but cannot fully capture its interaction with facial visual features, so its advantage largely disappears under domain shift. In contrast, MFGaze conditionally modulates visual features according to head pose, enabling more effective exploitation of head pose information and stronger cross-dataset generalization.

\subsubsection{\textbf{Comparison with State-of-the-Art Methods}}
Following the ablation studies, MFGaze was compared with mainstream gaze estimation methods on three public benchmarks to evaluate its competitiveness under standard evaluation protocols. All methods were trained and tested using the same experimental settings, and mean angular error (MAE) was adopted as the evaluation metric. As shown in Table~\ref{tab:sota_gaze_comparison}, MFGaze consistently achieved the lowest MAE across all three datasets, reaching 4.06$^\circ$ on MPIIFaceGaze, 4.23$^\circ$ on ETH-XGaze, and 10.52$^\circ$ on Gaze360.

Compared with conventional CNN-based models, the proposed method consistently achieved lower estimation errors, indicating better gaze estimation performance across datasets with varied acquisition conditions. MFGaze also outperformed GazeTR, the strongest Transformer-based baseline, by 0.22$^\circ$, 0.44$^\circ$, and 1.16$^\circ$ on MPIIFaceGaze, ETH-XGaze, and Gaze360, respectively. The largest improvement on Gaze360 suggests that the FiLM-based head-pose conditioning enhances robustness under unconstrained environments with substantial appearance variations.

\subsubsection{\textbf{Cross-Dataset Generalization Capability Testing}}
Real-world deployment inevitably involves variations in camera configurations, illumination conditions, and participant populations. To assess the ability of the proposed model to generalize across such distribution shifts, cross-dataset experiments were conducted among ETH-XGaze, MPIIFaceGaze, and Gaze360. In each transfer setting, the first dataset was used for training and the second for testing. 

As shown in Table~\ref{tab:cross_dataset_results}, conventional CNN-based methods exhibited substantial performance degradation under cross-dataset evaluation, indicating limited robustness to changes in acquisition environments and imaging conditions. Although GazeTR also adopts a Transformer architecture, its cross-dataset performance did not consistently surpass that of CNN-based models. In contrast, MFGaze achieved the lowest MAE in five of the six transfer settings. The only exception occurred in the MPIIFaceGaze→ETH-XGaze setting, where ResNet-18 achieved a marginally lower MAE than MFGaze (23.68$^\circ$ versus 23.85$^\circ$). Overall, the results consistently demonstrate that MFGaze provides superior cross-dataset robustness.

The transfer tasks between ETH-XGaze and Gaze360 present the largest domain gap in terms of head-pose distribution, illumination conditions, and acquisition environments. Compared with the second-best method, MFGaze reduced the MAE by 3.04$^\circ$ for the ETH-XGaze→Gaze360 transfer and by 3.14$^\circ$ for the Gaze360→ETH-XGaze transfer. This indicates that the FiLM-based head-pose conditioning enables the model to learn gaze representations that are less sensitive to domain-specific appearance variations than those learned by existing methods.

\subsection{Performance Evaluation of Gaze-to-Screen Calibration}

The polynomial ridge regression model described in Section III-C was adopted for gaze-to-screen calibration. Nine participants each provided 375 calibration samples, yielding 3,375 samples in total. In accordance with the individualized calibration setting, a separate model was fitted and evaluated for each participant. Five-fold cross-validation was performed within each participant, with samples grouped by calibration target so that all frames from the same target were assigned to the same fold. The reported errors were computed by pooling the out-of-fold predictions across the nine participants. The model used a second-order polynomial feature expansion, with the ridge regularization coefficient $\alpha=0.05$ and the peripheral weighting coefficient $\eta=0.6$. Pixel-level errors were converted into metric distances (mm) and angular errors ($^\circ$) based on the display geometry at a viewing distance of 1000 mm.

\begin{figure}[!t]
\centering
\subfloat[Spatial Error Map]{
\includegraphics[width=0.46\linewidth]{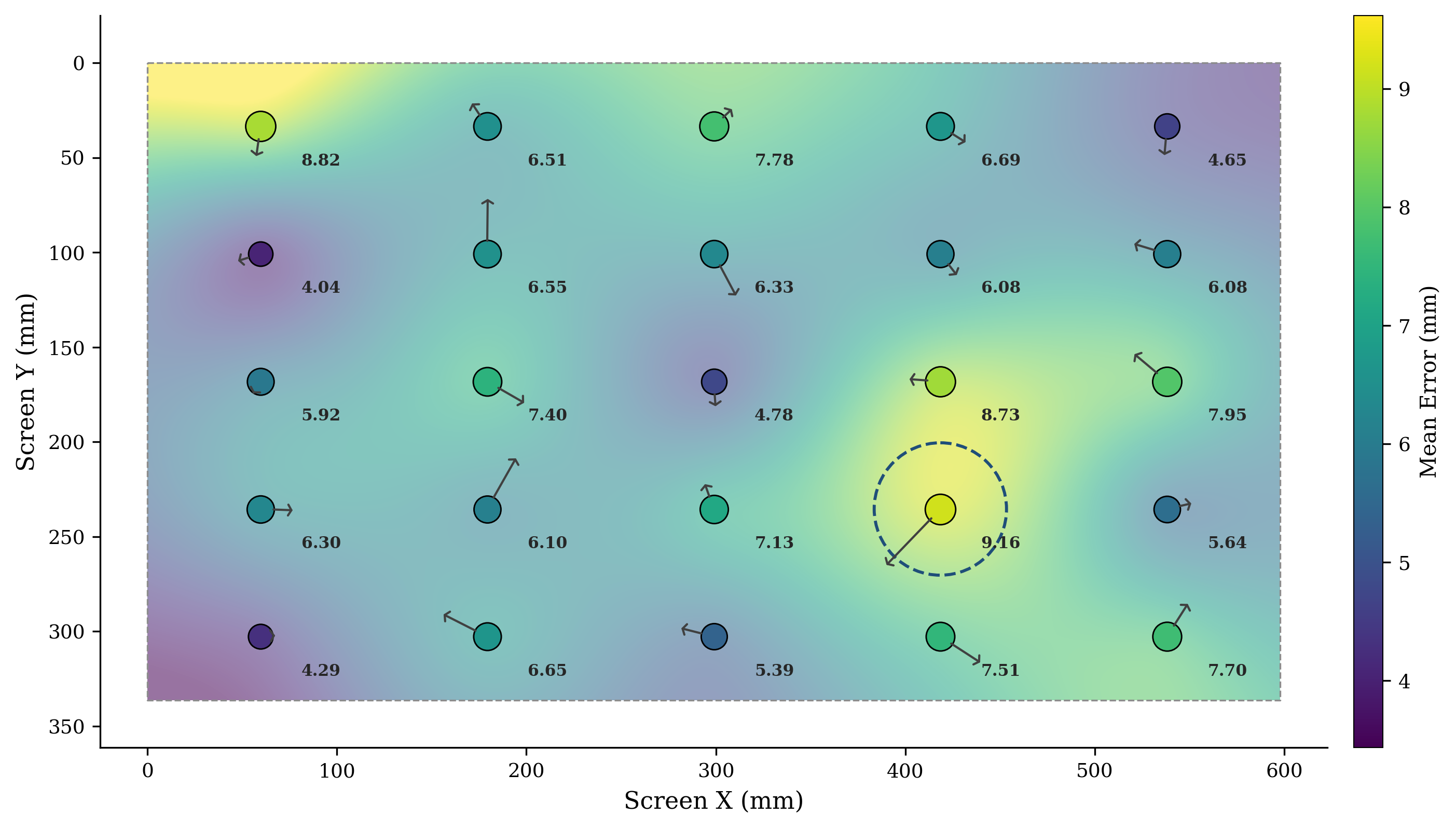}
\label{fig:calib_spatial_error}
}
\hfill
\subfloat[Error Distribution]{
\includegraphics[width=0.46\linewidth]{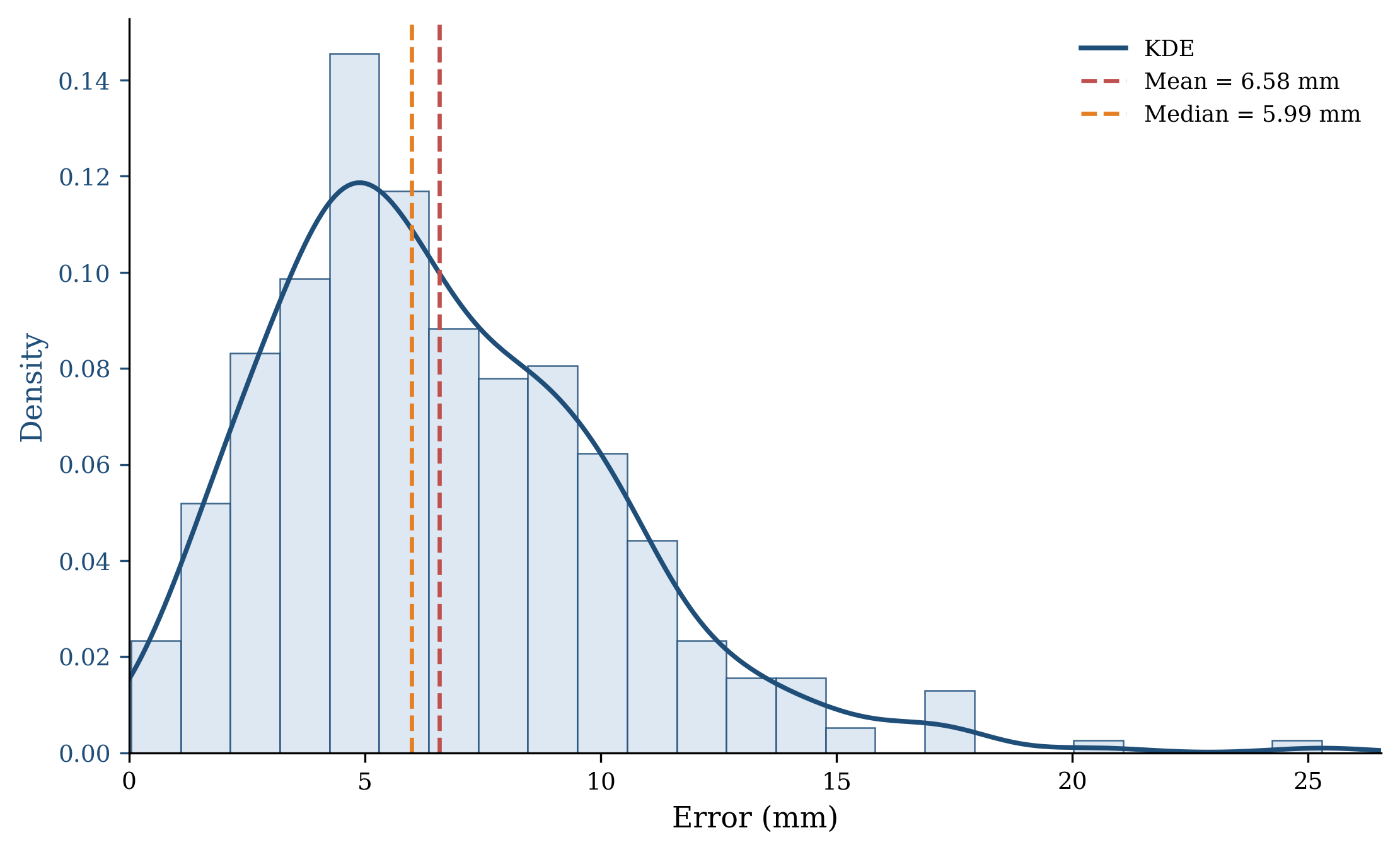}
\label{fig:calib_error_distribution}
}
\caption{Calibration error analysis of the gaze-to-screen calibration module. (a) Spatial distribution of the 25 calibration targets; marker color encodes the mean per-target error, and arrows (magnified ×6) indicate the mean direction of the predicted gaze offset. (b) Per-sample error distribution; the histogram and kernel density estimate are shown, with dashed lines marking the mean and median.}
\label{fig:calibration_error_analysis}
\end{figure}

Fig.~\ref{fig:calib_spatial_error} illustrates the spatial distribution of the 25 calibration targets and corresponding prediction errors. The results show no evident spatial bias or systematic error accumulation toward the screen boundaries, suggesting that the calibration model maintains stable performance across different screen locations.

Fig.~\ref{fig:calib_error_distribution} presents the error distribution across all samples. The proposed calibration method achieves a mean error of 6.58 mm (0.38$^\circ$) and a median error of 5.99 mm (0.34$^\circ$). The 95th-percentile error was 12.99 mm (0.74$^\circ$), and the maximum error was 25.28 mm (1.45$^\circ$); both remained within the approximate 2$^\circ$ foveal range. Overall, the calibration accuracy remains within the range of foveal vision tolerance, indicating that the estimated point of gaze provides sufficient spatial precision for downstream gaze target identification tasks.

\subsection{Performance Evaluation of Gaze Target Identification}

Gaze target identification is the final stage of the proposed framework and depends on the joint output of the gaze estimation, calibration, and scene understanding modules. This subsection evaluates the gaze target inference algorithm proposed in Section III-D on the end-to-end test dataset described in Section IV-C.

\begin{table*}[t]
    \centering
    \caption{Comparison of gaze target identification. Each method entry
    reports the number of correct predictions, with accuracy in parentheses.}
    \label{tab:gaze_target_identification}

    \small
    \setlength{\tabcolsep}{7pt}
    \renewcommand{\arraystretch}{1.10}

    \begin{tabular}{l c c c c c}
        \toprule
        \textbf{Class} & \textbf{N} & \textbf{M1} & \textbf{M2}
        & \textbf{M3} & \textbf{M4} \\
        \midrule

        Car
        & 142
        & 116 (81.69\%)
        & 106 (74.65\%)
        & 137 (96.48\%)
        & \textbf{142 (100\%)} \\

        Person
        & 118
        & 95 (80.51\%)
        & 79 (66.95\%)
        & 113 (95.76\%)
        & \textbf{115 (97.46\%)} \\

        Traffic sign
        & 89
        & 70 (78.65\%)
        & 61 (68.54\%)
        & \textbf{84 (94.38\%)}
        & \textbf{84 (94.38\%)} \\

        Traffic light
        & 76
        & 60 (78.95\%)
        & 51 (67.11\%)
        & 73 (96.05\%)
        & \textbf{74 (97.37\%)} \\

        Bicycle
        & 47
        & 38 (80.85\%)
        & 24 (51.06\%)
        & 44 (93.62\%)
        & \textbf{45 (95.74\%)} \\[1ex]

        \textbf{Task-relevant objects}
        & 472
        & 379 (80.30\%)
        & 321 (68.01\%)
        & 451 (95.55\%)
        & \textbf{460 (97.46\%)} \\

        \midrule

        Building
        & 96
        & 65 (67.71\%)
        & 79 (82.29\%)
        & 83 (86.46\%)
        & \textbf{89 (92.71\%)} \\

        Road
        & 78
        & 56 (71.79\%)
        & 67 (85.90\%)
        & 70 (89.74\%)
        & \textbf{75 (96.15\%)} \\

        Sidewalk
        & 52
        & 35 (67.31\%)
        & 42 (80.77\%)
        & 44 (84.62\%)
        & \textbf{49 (94.23\%)} \\

        Vegetation
        & 64
        & 43 (67.19\%)
        & 53 (82.81\%)
        & 55 (85.94\%)
        & \textbf{60 (93.75\%)} \\

        Sky
        & 42
        & 31 (73.81\%)
        & 35 (83.33\%)
        & 37 (88.10\%)
        & \textbf{41 (97.62\%)} \\[1ex]

        \textbf{Background regions}
        & 332
        & 230 (69.28\%)
        & 276 (83.13\%)
        & 289 (87.05\%)
        & \textbf{314 (94.58\%)} \\

        \midrule

        \textbf{overall}
        & 804
        & 609 (75.75\%)
        & 597 (74.25\%)
        & 740 (92.04\%)
        & \textbf{774 (96.27\%)} \\

        \bottomrule
    \end{tabular}
\end{table*}

\subsubsection{\textbf{Ablation Studies}}

The dataset covers ten classes of candidate objects from the Cityscapes test images, organized by semantic class and divided into two groups: task-relevant objects (car, person, traffic sign, traffic light, and bicycle) and background regions (road, sidewalk, building, vegetation, sky). The test set contained 804 valid samples, and the evaluation metric was Top-1 identification accuracy.

To quantify the independent contributions of each design component in Section III-D, four method configurations were constructed as baseline comparisons. M1 used pure centroid distance matching, taking the Euclidean distance from the gaze point to the geometric centroid of the candidate object as the sole criterion; M2 used the shortest distance from the gaze point to the edge of the candidate object's mask. M3 introduced class-specific semantic weights on top of M2 but fixed the fusion coefficient $\lambda$ at 1.0. M4 was the complete method, which further incorporated the entropy-driven adaptive fusion mechanism proposed in Section III-D on top of M3, allowing $\lambda$ to adapt within the range $[\lambda_{min}, \lambda_{max}]$ according to the degree of scene ambiguity. 

The decay scale parameter $\tau$ was set to 80 px, corresponding to approximately 21.79 mm on the screen. This value lay between the calibration error of 24 px (the 6.58 mm reported in Section V-B) and the human foveal tolerance of approximately 128 px under a 2$^\circ$ visual angle, so that the identification decisions were aligned with the natural fixation precision of human observers. The fusion coefficient bounds $\lambda_{min}$ and $\lambda_{max}$ were set to 0.2 and 1.0, respectively, and the semantic weights for the 10 classes in Cityscapes followed the values in Table~\ref{tab:semantic_weights}. 

As shown in Table~\ref{tab:gaze_target_identification}, M2 exhibits contrasting performance across the two object groups. Compared with M1, M2 improves the accuracy of background regions but reduces the accuracy of task-relevant objects. This is because boundary-distance matching is better suited to large or irregular background regions, whose centroids may be far from the gaze point even when the point falls within the region. However, for small task-relevant objects adjacent to large background regions, slight gaze offsets may cause the target object and the neighboring background region to receive comparable geometric scores, leading to incorrect assignments without semantic guidance.

Incorporating class-level semantic priors in M3 effectively mitigates this ambiguity. Task-relevant objects such as vehicles and pedestrians are assigned higher prior weights, reflecting empirically observed driver fixation preferences. As a result, M3 substantially improves the identification performance of task-relevant objects, leading to a significant increase in overall accuracy.

M4 further introduces an entropy-driven adaptive fusion mechanism that dynamically adjusts the contribution of semantic priors based on geometric ambiguity. When the scene contains sparse and well-separated objects, geometric evidence dominates the decision process; when multiple candidates are densely clustered, semantic priors play a stronger role in disambiguation. This adaptive behavior yields the highest overall accuracy of 96.27\%, exceeding M3 by 4.23 percentage points.

\subsubsection{\textbf{Hyperparameter Sensitivity Analysis}}

The proposed method involves several empirically defined hyperparameters, including the decay scale parameter $\tau$, the fusion coefficient bounds $\lambda_{min}$ and $\lambda_{max}$, and the class-level semantic weights for the Cityscapes classes. The specific values of these parameters involve a degree of design choice. To assess the robustness of identification accuracy to small perturbations, sensitivity analyses were conducted on these three key types of hyperparameters.

The parameter $\tau$ controls how fast the geometric matching score decays with boundary distance; it characterizes the distance within which a gaze deviation can still be treated as a fixation on the target. In this experiment, five candidate values were selected between 50\% and 200\% of the optimal value. All other hyperparameters were fixed, and the evaluation was rerun on the test set. The results are shown in Table~\ref{tab:sensitivity_tau}.

\begin{table}[!htbp]
    \centering
    \caption{Sensitivity analysis of the decay scale parameter $\tau$}
    \label{tab:sensitivity_tau}
    \resizebox{\columnwidth}{!}{
    \begin{tabular}{l c c c c c}
        \toprule
        $\tau$ (relative to the optimum) & 50\% & 75\% & 100\% & 150\% & 200\% \\
        \midrule
        Actual value of $\tau$ & 40 px & 60 px & 80 px & 120 px & 160 px \\
        Overall accuracy & 95.77\% & 95.90\% & \textbf{96.27\%} & 95.90\% & 95.52\% \\
        \bottomrule
    \end{tabular}
    }
\end{table}
\FloatBarrier

Over the tested range of $\tau$, the overall identification accuracy varied between 95.52\% and 96.27\%, a maximum fluctuation of only 0.75 pp, indicating that identification is robust to the specific value of $\tau$. Accuracy was highest at $\tau=80$ px and declined slightly toward both ends of the range. When $\tau$ was too small, such as 40 px, the geometric matching score decayed too rapidly, causing some true gaze targets to be underestimated when their boundary distances were slightly larger, reducing accuracy to 95.77\%. When $\tau$ was too large, such as 160 px, the score decayed too slowly, allowing non-target candidates far from the gaze point to receive spuriously high geometric scores, reducing accuracy to 95.52\%. The slight decline at both ends supports the choice of $\tau=80$ px.

\begin{table}[!htbp]
    \centering
    \caption{Sensitivity analysis of the fusion coefficient bounds}
    \label{tab:sensitivity_bounds}
    \begin{tabular}{l c c c}
        \toprule
        Configuration & $\lambda_{min}$ & $\lambda_{max}$ & Overall accuracy \\
        \midrule
        \textbf{C1 (ours)} & 0.2 & 1.0 & \textbf{96.27\%} \\
        C2 & 0.0 & 1.0 & 95.15\% \\
        C3 & 0.4 & 1.0 & 96.02\% \\
        C4 & 0.2 & 0.8 & 95.40\% \\
        C5 & 0.2 & 1.2 & 96.14\% \\
        \bottomrule
    \end{tabular}
\end{table}
\FloatBarrier

As shown in Table~\ref{tab:sensitivity_bounds}, the overall accuracy under the five configurations C1--C5 remained within 95.15\% to 96.27\%, a maximum fluctuation of only 1.12 pp. This indicates that the bound setting does not depend on a precise numerical choice; the adaptive fusion mechanism tolerated reasonable variations in the bounds well.

\begin{table}[!htbp]
    \centering
    \setlength{\tabcolsep}{3pt}
    \caption{Perturbation test of class-level semantic weights}
    \label{tab:perturbation_weights}
    \begin{tabular}{l c c}
        \toprule
        Perturbation magnitude
        & \textbf{Average overall accuracy}
        & \makecell{\textbf{Standard}\\\textbf{deviation (\%)}} \\
        \midrule
        $\pm$ 0.0 (Ours) & \textbf{96.27\%} & -- \\
        $\pm$ 0.1 & 96.01\% & 0.32 \\
        $\pm$ 0.2 & 95.64\% & 0.61 \\
        \bottomrule
    \end{tabular}
\end{table}
\FloatBarrier

As shown in Table~\ref{tab:perturbation_weights}, under random perturbations of $\pm 0.1$ and $\pm 0.2$, the average overall accuracies over 10 runs were 96.01\% and 95.64\%, respectively. Compared with the original weights, the accuracy decrease was no greater than 0.63 pp. The corresponding standard deviations were 0.32\% and 0.61\%, indicating limited variation across repeated runs. These results show that identification is also robust to the specific values of the class-level semantic weights.

Overall, these sensitivity analyses show that the proposed method maintained stable performance under reasonable perturbations of empirically defined hyperparameters.

\FloatBarrier
\subsection{Performance Evaluation on the LBW Dataset}

The proposed framework was further evaluated on the LBW dataset after sequence-specific calibration following the protocol described in Section IV-D. The same M1--M4 configurations and evaluation criterion as those used in Section V-C were adopted. Considering the lower resolution of the LBW forward-view images, the pixel-based decay scale $\tau$ was adjusted in proportion to image width to preserve the normalized spatial tolerance used in Section V-C. Accordingly, $\tau$ was set to 34 pixels, while the remaining inference parameters were kept unchanged. Under the reporting scope defined in Section IV-D, 55,550 evaluation frames were included, comprising 48,015 road targets and 7,535 task-relevant object targets. Table~\ref{tab:lbw_target_identification} presents the class-wise Top-1 gaze target identification accuracy for task-relevant objects and the road class, together with the pooled accuracy across all task-relevant object classes.

\begin{table}[!htbp]
    \centering
    \caption{Top-1 gaze target identification accuracy for road and task-relevant objects on the LBW dataset}
    \label{tab:lbw_target_identification}

    \small
    \setlength{\tabcolsep}{4pt}
    \renewcommand{\arraystretch}{1.08}

    \begin{tabular}{lccccc}
        \toprule
        Target class & $N$ & M1 & M2 & M3 & M4 \\
        \midrule

        road
        & 48,015
        & 56.23\%
        & 83.37\%
        & 83.61\%
        & \textbf{85.58\%} \\

        \midrule

        car
        & 7,201
        & 72.18\%
        & 63.69\%
        & 78.27\%
        & \textbf{81.13\%} \\

        bicycle
        & 19
        & 68.42\%
        & 31.58\%
        & \textbf{73.68\%}
        & \textbf{73.68\%} \\

        person
        & 101
        & 53.47\%
        & 27.72\%
        & 50.50\%
        & \textbf{64.36\%} \\

        traffic light
        & 119
        & 15.97\%
        & 3.36\%
        & 38.66\%
        & \textbf{47.06\%} \\

        traffic sign
        & 95
        & 28.42\%
        & 8.42\%
        & 38.95\%
        & \textbf{47.37\%} \\

        \shortstack[l]{\textbf{Task-relevant}\\\textbf{objects}}
        & \textbf{7,535}
        & 70.48\%
        & 61.47\%
        & 76.76\%
        & \textbf{79.92\%} \\

        \midrule

        \textbf{Overall}
        & \textbf{55,550}
        & 58.17\%
        & 80.40\%
        & 82.68\%
        & \textbf{84.82\%} \\

        \bottomrule
    \end{tabular}
\end{table}
\FloatBarrier

As shown in Table~\ref{tab:lbw_target_identification}, M4 achieved Top-1 gaze target identification accuracies of 79.92\% for task-relevant objects and 85.58\% for the road class. The comparison among M1--M4 demonstrates the contribution of boundary-distance geometry and semantic priors to target identification under naturalistic driving conditions. Compared with M3, M4 improves the pooled accuracy for task-relevant objects by 3.16 percentage points.

The class imbalance in LBW reflects both the composition of natural driving scenes and drivers' gaze behavior: road regions and vehicles are more prevalent in the forward view and are also attended to more frequently than pedestrians, bicycles, traffic signs, and traffic lights. The lower accuracies for traffic signs and traffic lights may partly reflect the relatively low resolution of the LBW forward-view images: these targets, especially when distant, occupy only small image regions and are therefore particularly sensitive to minor gaze-to-scene mapping errors. Overall, the results demonstrate the effectiveness of the proposed pipeline for gaze target identification under naturalistic driving conditions.

\section{Conclusion}

This paper has presented a modular framework for identifying the object-level gaze target attended to by the driver from synchronized in-cabin and forward-facing images. The proposed framework moves beyond prior formulations based on gaze-zone classification and continuous spatial attention localization. Through the decoupling of gaze estimation, scene understanding, gaze-to-scene calibration and mapping, and gaze target inference, each sub-module can be designed, evaluated, and diagnosed independently, while the overall system operates under a low-cost camera-based configuration. The inferred gaze targets provide object-level information for driver visual attention monitoring.

The experimental results validate the proposed framework at three levels. First, MFGaze, which combines an MAE-pretrained ViT backbone with FiLM-based head-pose modulation, achieves the lowest gaze estimation error among the compared methods on three public datasets and demonstrates strong cross-dataset generalization. Second, the polynomial calibration module keeps 95\% of predicted gaze points in the simulator experiment within the approximate range of human foveal vision, providing sufficient geometric precision for object-level reasoning. Third, the fusion of boundary-distance geometry and class-level semantic
priors achieves an end-to-end Top-1 gaze target identification accuracy of 96.27\% in the laboratory experiment, with particularly large gains for task-relevant objects. Evaluation
on the LBW dataset after sequence-specific calibration further shows that M4 achieves Top-1 gaze target identification accuracies of 79.92\% for task-relevant objects and 85.58\% for the road class. The sensitivity analyses further show that the proposed method maintains stable performance under perturbations of empirically defined hyperparameters.

Several limitations also point to future work. The current framework
performs target reasoning in the two-dimensional image plane and does not explicitly incorporate depth information into gaze target inference. Extending the inference stage from image-plane reasoning to three-dimensional scene reasoning may help disambiguate near and far objects under severe occlusion. In addition, the current framework operates on individual frames, whereas driver gaze behavior has strong temporal structure in the form of fixations, saccades, and target transitions. Incorporating sequence-level reasoning over the inferred target stream is expected to improve robustness against transient errors in upstream modules. The framework also relies on participant-specific calibration in the simulator and sequence-specific calibration in LBW. Calibration stability under changes in camera configuration remains to be evaluated.


%





\ifCLASSOPTIONcaptionsoff
  \newpage
\fi




\begin{thebibliography}{10}

\bibitem{1}
P. K. Sharma and P. Chakraborty, ``A review of driver gaze estimation and application in gaze behavior understanding,'' \emph{Eng. Appl. Artif. Intell.}, vol. 133, Art. no. 108117, 2024.

\bibitem{2}
S. G. Klauer, T. A. Dingus, V. L. Neale, J. D. Sudweeks, and D. J. Ramsey, ``The impact of driver inattention on near-crash/crash risk: An analysis using the 100-Car Naturalistic Driving Study data,'' National Highway Traffic Safety Administration, Washington, DC, USA, Rep. DOT HS 810 594, Apr. 2006.

\bibitem{3}
I. Kotseruba and J. K. Tsotsos, ``Attention for vision-based assistive and automated driving: A review of algorithms and datasets,'' \emph{IEEE Trans. Intell. Transp. Syst.}, vol. 23, no. 11, pp. 19907--19928, 2022.

\bibitem{4}
SAE International, ``Taxonomy and definitions for terms related to driving automation systems for on-road motor vehicles,'' SAE Standard J3016\_202104, Apr. 2021.

\bibitem{5}
National Center for Statistics and Analysis, ``Distracted driving in 2022,'' National Highway Traffic Safety Administration, Washington, DC, USA, Traffic Safety Facts Research Note, Rep. DOT HS 813 559, Apr. 2024.

\bibitem{6}
S. Vora, A. Rangesh, and M. M. Trivedi, ``Driver gaze zone estimation using convolutional neural networks: A general framework and ablative analysis,'' \emph{IEEE Trans. Intell. Veh.}, vol. 3, no. 3, pp. 254--265, 2018.

\bibitem{7}
A. Jain, H. S. Koppula, S. Soh, B. Raghavan, A. Singh, and A. Saxena, ``Brain4Cars: Car that knows before you do via sensory-fusion deep learning architecture,'' arXiv:1601.00740, 2016.

\bibitem{8}
K. Holmqvist, M. Nystr\"{o}m, R. Andersson, R. Dewhurst, H. Jarodzka, and J. van de Weijer, \emph{Eye Tracking: A Comprehensive Guide to Methods and Measures}. Oxford, U.K.: Oxford Univ. Press, 2011.

\bibitem{9}
Y. Cheng, H. Wang, Y. Bao, and F. Lu, ``Appearance-based gaze estimation with deep learning: A review and benchmark,'' \emph{IEEE Trans. Pattern Anal. Mach. Intell.}, vol. 46, no. 12, pp. 7509--7528, Dec. 2024.

\bibitem{10}
L. Fridman, J. Lee, B. Reimer, and T. Victor, ``'Owl' and 'Lizard': Patterns of head pose and eye pose in driver gaze classification,'' \emph{IET Comput. Vis.}, vol. 10, no. 4, pp. 308--314, 2016.

\bibitem{11}
A. Tawari and M. M. Trivedi, ``Robust and continuous estimation of driver gaze zone by dynamic analysis of multiple face videos,'' in \emph{Proc. IEEE Intell. Veh. Symp. (IV)}, 2014, pp. 344--349.

\bibitem{12}
Y. Zhang, X. Yang, and Z. Ma, ``Driver's gaze zone estimation method: A four-channel convolutional neural network model,'' in \emph{Proc. 2nd Int. Conf. Big Data Service Intell. Comput.}, 2020, pp. 20--24.

\bibitem{13}
A. Palazzi, D. Abati, S. Calderara, F. Solera, and R. Cucchiara, ``Predicting the driver's focus of attention: The DR(eye)VE project,'' \emph{IEEE Trans. Pattern Anal. Mach. Intell.}, vol. 41, no. 7, pp. 1720--1733, 2019.

\bibitem{14}
C. Xu, B. Jiang, Y. Wang, and Y. Su, ``A modeled study of driver visual attention driven by driving tasks,'' \emph{Eng. Appl. Artif. Intell.}, vol. 157, Art. no. 111382, Oct. 2025.

\bibitem{15}
Z. Hu, C. Lv, P. Hang, C. Huang, and Y. Xing, ``Data-driven estimation of driver attention using calibration-free eye gaze and scene features,'' \emph{IEEE Trans. Ind. Electron.}, vol. 69, no. 2, pp. 1800--1808, 2022.

\bibitem{Kasahara2022LBW}
I. Kasahara, S. Stent, and H. S. Park, ``Look both ways: Self-supervising driver gaze estimation and road scene saliency,'' in \emph{Proc. Eur. Conf. Comput. Vis. (ECCV)}, 2022, pp. 126--142.

\bibitem{16}
Z. Zhu and Q. Ji, ``Novel eye gaze tracking techniques under natural head movement,'' \emph{IEEE Trans. Biomed. Eng.}, vol. 54, no. 12, pp. 2246--2260, 2007.

\bibitem{17}
E. D. Guestrin and M. Eizenman, ``General theory of remote gaze estimation using the pupil center and corneal reflections,'' \emph{IEEE Trans. Biomed. Eng.}, vol. 53, no. 6, pp. 1124--1133, 2006.

\bibitem{18}
J. Zhang, Y. Zhou, R. Xi, et al., ``EyeGaze: Hybrid eye tracking for authentication on mobile devices,'' \emph{Proc. ACM Interact. Mobile Wearable Ubiquitous Technol.}, vol. 7, no. 2, Art. no. 64, 2023.

\bibitem{19}
D. W. Hansen and Q. Ji, ``In the eye of the beholder: A survey of models for eyes and gaze,'' \emph{IEEE Trans. Pattern Anal. Mach. Intell.}, vol. 32, no. 3, pp. 478--500, 2010.

\bibitem{20}
X. Zhang, S. Park, T. Beeler, D. Bradley, S. Tang, and O. Hilliges, ``ETH-XGaze: A large-scale dataset for gaze estimation under extreme head pose and gaze variation,'' in \emph{Proc. Eur. Conf. Comput. Vis. (ECCV)}, 2020, pp. 365--381.

\bibitem{21}
X. Zhang, Y. Sugano, M. Fritz, and A. Bulling, ``MPIIGaze: Real-world dataset and deep appearance-based gaze estimation,'' \emph{IEEE Trans. Pattern Anal. Mach. Intell.}, vol. 41, no. 1, pp. 162--175, 2019.

\bibitem{22}
P. Kellnhofer, A. Recasens, S. Stent, W. Matusik, and A. Torralba, ``Gaze360: Physically unconstrained gaze estimation in the wild,'' in \emph{Proc. IEEE/CVF Int. Conf. Comput. Vis. (ICCV)}, 2019, pp. 6912--6921.

\bibitem{23}
X. Zhang, Y. Sugano, M. Fritz, and A. Bulling, ``It's written all over your face: Full-face appearance-based gaze estimation,'' in \emph{Proc. IEEE Conf. Comput. Vis. Pattern Recognit. Workshops (CVPRW)}, 2017, pp. 51--60.

\bibitem{24}
T. Fischer, H. J. Chang, and Y. Demiris, ``RT-GENE: Real-time eye gaze estimation in natural environments,'' in \emph{Proc. Eur. Conf. Comput. Vis. (ECCV)}, 2018, pp. 339--357.

\bibitem{25}
A. Dosovitskiy et al., ``An image is worth 16 $\times$ 16 words: Transformers for image recognition at scale,'' in \emph{Proc. Int. Conf. Learn. Represent. (ICLR)}, 2021.

\bibitem{26}
Y. Cheng and F. Lu, ``Gaze estimation using transformer,'' in \emph{Proc. 26th Int. Conf. Pattern Recognit. (ICPR)}, 2022, pp. 3341--3347.

\bibitem{27}
R. Karmi, R. Mastouri, I. Rahmany, and N. Khlifa, ``An appearance-based vision transformer network for enhanced gaze estimation,'' \emph{Signal Image Video Process.}, vol. 19, no. 9, Art. no. 742, 2025.

\bibitem{28}
D. Hu, M. Cui, and K. Huang, ``FIFA: Fine-grained inter-frame attention for driver's video gaze estimation,'' in \emph{Proc. IEEE/CVF Conf. Comput. Vis. Pattern Recognit. (CVPR)}, 2025, pp. 18760--18769.

\bibitem{31}
S. Ren, K. He, R. Girshick, and J. Sun, ``Faster R-CNN: Towards real-time object detection with region proposal networks,'' \emph{IEEE Trans. Pattern Anal. Mach. Intell.}, vol. 39, no. 6, pp. 1137--1149, 2017.

\bibitem{32}
N. Carion et al., ``End-to-end object detection with transformers,'' in \emph{Proc. Eur. Conf. Comput. Vis. (ECCV)}, 2020, pp. 213--229.

\bibitem{33}
J. Long, E. Shelhamer, and T. Darrell, ``Fully convolutional networks for semantic segmentation,'' \emph{IEEE Trans. Pattern Anal. Mach. Intell.}, vol. 39, no. 4, pp. 640--651, 2017.

\bibitem{34}
L.-C. Chen, Y. Zhu, G. Papandreou, F. Schroff, and H. Adam, ``Encoder-decoder with atrous separable convolution for semantic image segmentation,'' in \emph{Proc. Eur. Conf. Comput. Vis. (ECCV)}, 2018, pp. 801--818.

\bibitem{35}
K. He, G. Gkioxari, P. Doll\'{a}r, and R. Girshick, ``Mask R-CNN,'' \emph{IEEE Trans. Pattern Anal. Mach. Intell.}, vol. 42, no. 2, pp. 386--397, 2020.

\bibitem{36}
A. Kirillov, K. He, R. Girshick, C. Rother, and P. Doll\'{a}r, ``Panoptic segmentation,'' in \emph{Proc. IEEE/CVF Conf. Comput. Vis. Pattern Recognit. (CVPR)}, 2019, pp. 9404--9413.

\bibitem{37}
M. Cordts et al., ``The Cityscapes dataset for semantic urban scene understanding,'' in \emph{Proc. IEEE Conf. Comput. Vis. Pattern Recognit. (CVPR)}, 2016, pp. 3213--3223.

\bibitem{38}
A. Recasens, C. Vondrick, A. Khosla, and A. Torralba, ``Following gaze in video,'' in \emph{Proc. IEEE Int. Conf. Comput. Vis. (ICCV)}, 2017, pp. 1435--1443.

\bibitem{39}
I. Dua, T. A. John, R. Gupta, and C. Jawahar, ``DGAZE: Driver gaze mapping on road,'' in \emph{Proc. IEEE/RSJ Int. Conf. Intell. Robots Syst. (IROS)}, 2020, pp. 5946--5953.

\bibitem{40}
J. Araluce, L. M. Bergasa, M. Oca\~{n}a, et al., ``Gaze focalization system for driving applications using OpenFace 2.0 toolkit with NARMAX algorithm in accidental scenarios,'' \emph{Sensors}, vol. 21, no. 18, Art. no. 6262, 2021.

\bibitem{41}
J. Stapel, M. El Hassnaoui, and R. Happee, ``Measuring driver perception: Combining eye-tracking and automated road scene perception,'' \emph{Hum. Factors}, vol. 64, no. 4, pp. 714--731, 2022.

\bibitem{42}
M. Karuppasamy, S. Gangisetty, S. N. Rai, C. Masone, and C. V. Jawahar, ``Towards safer and understandable driver intention prediction,'' in \emph{Proc. IEEE/CVF Int. Conf. Comput. Vis. (ICCV)}, 2025, pp. 25378--25387.

\bibitem{43}
J. Zhou et al., ``EraW-Net: Enhance-refine-align W-Net for scene-associated driver attention estimation,'' \emph{IEEE Trans. Multimedia}, vol. 27, pp. 5922--5935, 2025.

\end{thebibliography}
%

\begin{IEEEbiography}[{\includegraphics[width=1in,height=1.25in,clip,keepaspectratio]{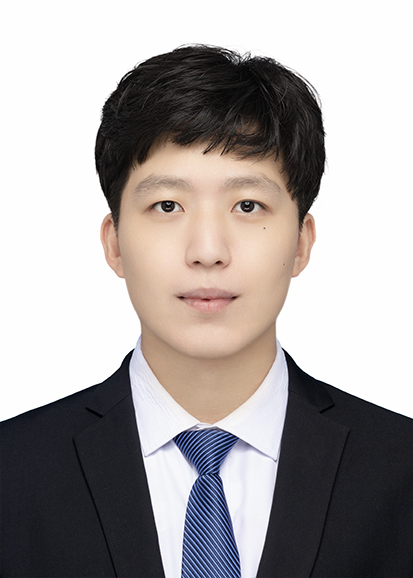}}]{Xinze Cao}
received the B.E. degree from Shandong University of Science and Technology, Qingdao, China, in 2024, where he is currently pursuing the M.S. degree with the College of Transportation. His research interests include driver gaze estimation, driver visual attention modeling, driver behavior analysis, driver state monitoring, and traffic safety evaluation.
\end{IEEEbiography}

\begin{IEEEbiography}[{\includegraphics[width=1in,height=1.25in,clip,keepaspectratio]{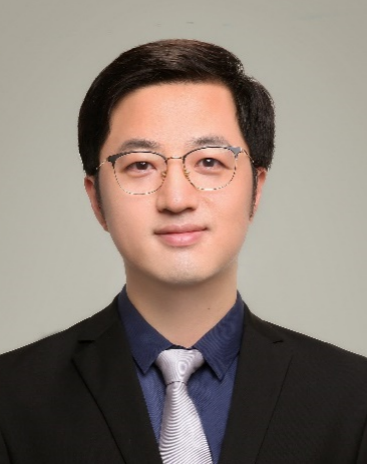}}]{Ke Wang}
received the B.Tech. degree from the Harbin Institute of Technology, Harbin, China, in 2008, M.S. degree from the Tongji University, Shanghai, China, in 2011, and the Ph.D. degree from University of California, Irvine, USA, in 2017. He is currently a Professor with the College of Transportation, Shandong University of Science and Technology. His research interests include driving behavior analysis, driver state monitoring, and traffic safety evaluation.
\end{IEEEbiography}

\begin{IEEEbiography}[{\includegraphics[width=1in,height=1.25in,clip,keepaspectratio]{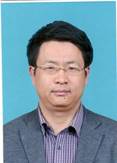}}]{Deyong Guan}
received the B.E. degree in automotive engineering from Shandong Institute of Engineering, China, in 1998, and the Ph.D. degree in transportation planning and management from Jilin University, Changchun, China, in 2003. From 2003 to 2005, he conducted postdoctoral research at Tongji University, Shanghai, China. He is currently a Professor with the College of Transportation, Shandong University of Science and Technology, Qingdao, China. His research interests include intelligent transportation systems, traffic management and control, transportation data analytics, and smart cities.
\end{IEEEbiography}

\begin{IEEEbiography}[{\includegraphics[width=1in,height=1.25in,clip,keepaspectratio]{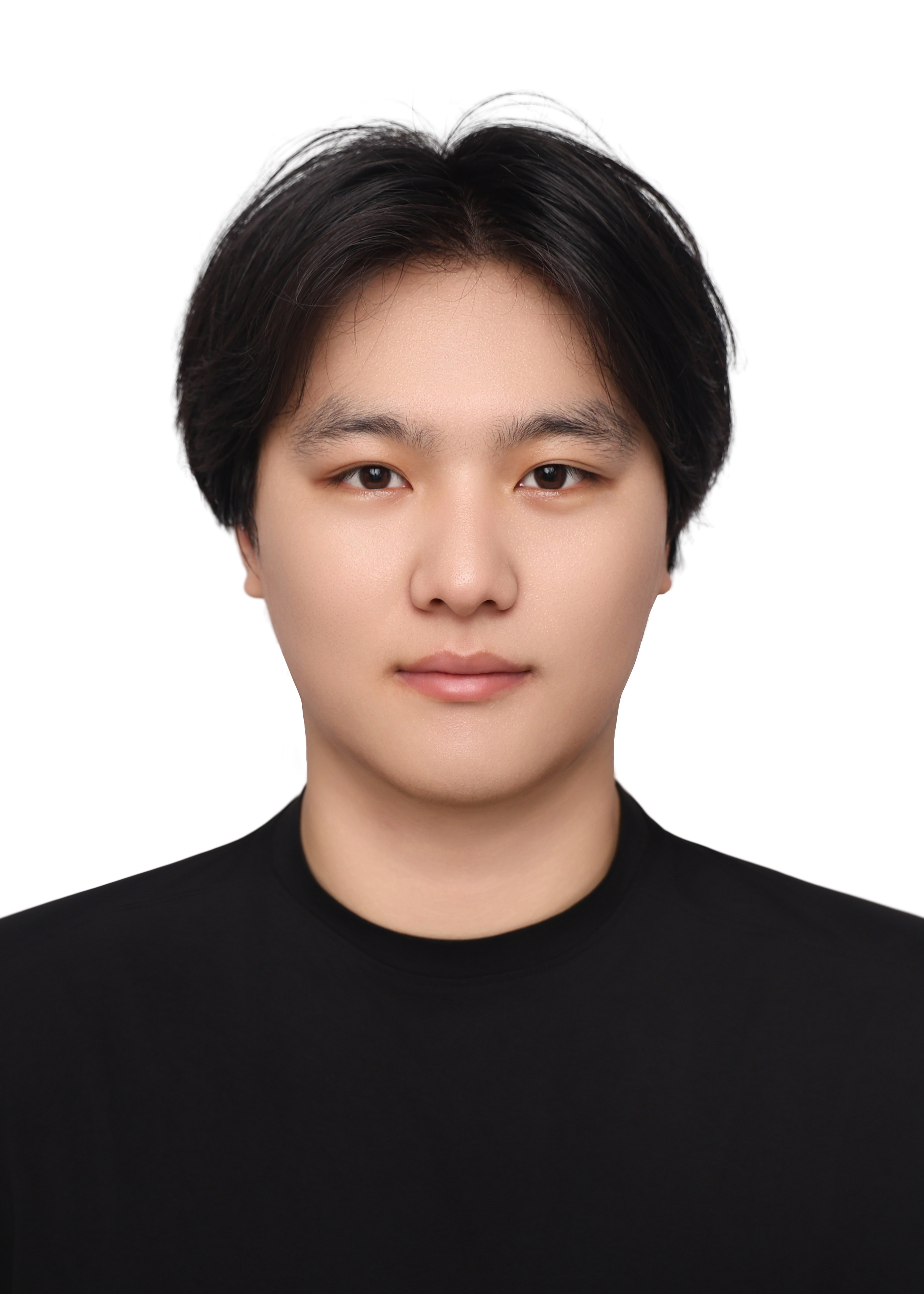}}]{Yuanquan Li}
received the B.E. degree from Liaocheng University, Liaocheng, China, in 2024. He is currently pursuing the M.S. degree with the College of Transportation, Shandong University of Science and Technology, Qingdao, China. His research interests include driver gaze estimation, driver visual attention modeling, driver behavior analysis, driver state monitoring, intelligent vehicles, and traffic safety.
\end{IEEEbiography}

\begin{IEEEbiography}[{\includegraphics[width=1in,height=1.25in,clip,keepaspectratio]{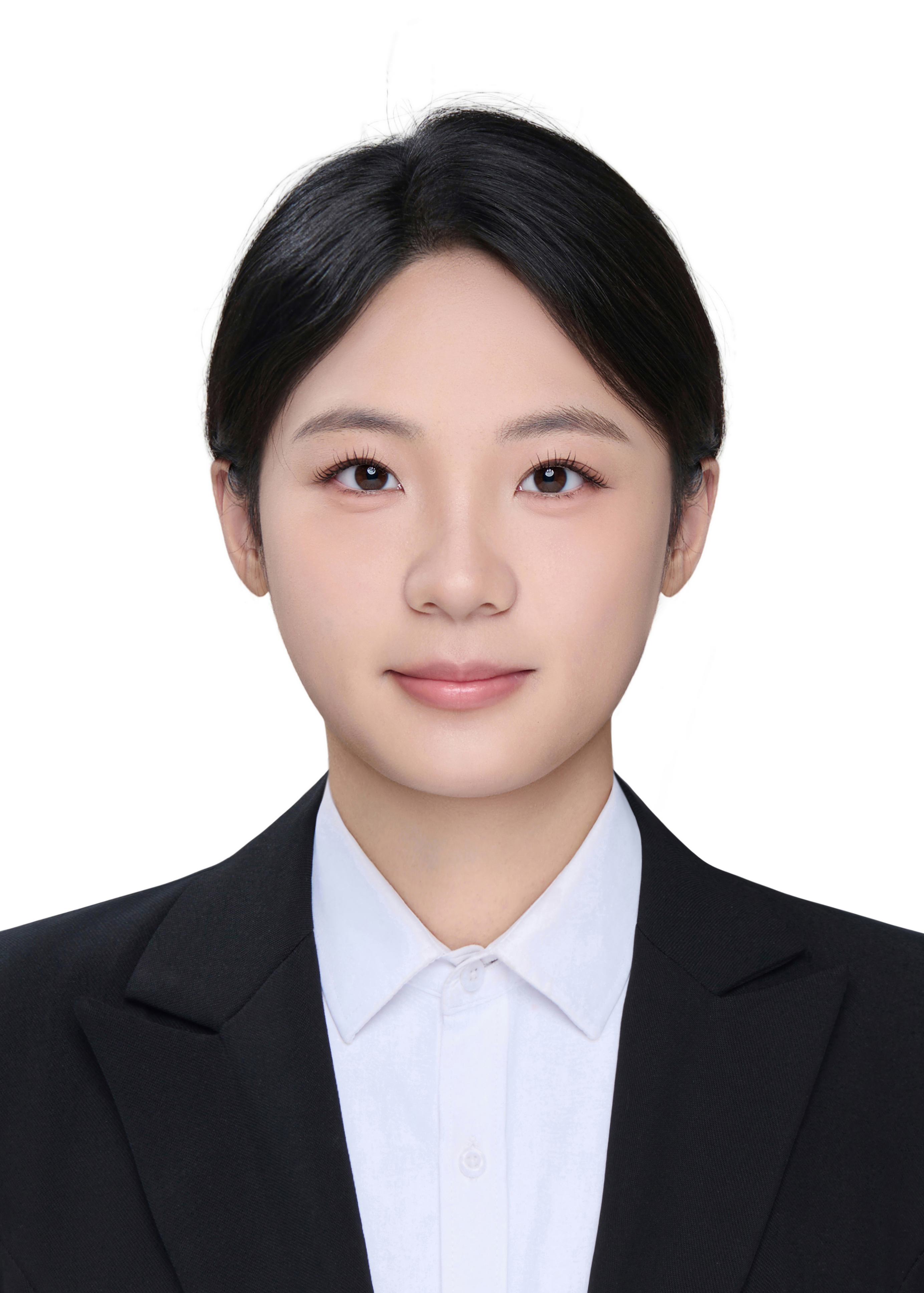}}]{Ke Zhang}
received the B.E. degree from Shandong Jiaotong University, Jinan, China. She has been pursuing the M.S. degree with the College of Transportation, Shandong University of Science and Technology, Qingdao, China, since 2024. Her research interests include traffic flow analysis, intelligent driving, and driver behavior analysis.
\end{IEEEbiography}



\vfill


\end{document}